\documentclass[journal,letterpaper,twoside]{IEEEtran}

\usepackage[cmex10]{amsmath}
\usepackage{latexsym,amssymb,cite,bbm,epsf,amsthm,subfigure,algorithm,algorithmic}
\usepackage{graphicx}
\usepackage{enumerate}
\usepackage{mathrsfs}
\usepackage{epstopdf}
\usepackage{bbm}
\usepackage{color}
\usepackage{url}
\usepackage{threeparttable}

\newcommand{\bc}{\begin{center}}
\newcommand{\ec}{\end{center}}
\newcommand{\be}{\begin{equation}}
\newcommand{\ee}{\end{equation}}
\newcommand{\ba}{\begin{array}}
\newcommand{\ea}{\end{array}}
\newcommand{\bea}{\begin{eqnarray}}
\newcommand{\eea}{\end{eqnarray}}
\newcommand{\bal}{\begin{align}}
\newcommand{\eal}{\end{align}}
\newcommand{\ei}{\end{itemize}}
\newcommand{\bi}{\begin{itemize}}
\newcommand{\bfi}{\begin{figure}}
\newcommand{\efi}{\end{figure}}
\newcommand{\MB}{\left[\begin{array}}
\newcommand{\ME}{\end{array}\right]}
\newcommand{\nn}{\nonumber}

\renewcommand{\vec}[1]{\mbox{\boldmath${#1}$}}
\newcommand{\mA}{\vec{A}}
\newcommand{\mX}{\vec{X}}
\newcommand{\mY}{\vec{Y}}
\newcommand{\mZ}{\vec{Z}}
\newcommand{\mZt}{\overset{\sim}{\vec{Z}}}
\newcommand{\mH}{\vec{H}}
\newcommand{\mI}{\vec{I}}

\newcommand{\mU}{\vec{U}}
\newcommand{\mV}{\vec{V}}

\newcommand{\mE}{\vec{E}}

\newcommand{\mSig}{\vec{\Sigma}}
\newcommand{\mOm}{\vec{\Omega}}
\newcommand{\mB}{\vec{B}}
\newcommand{\mC}{\vec{C}}
\newcommand{\mF}{\vec{F}}
\newcommand{\mG}{\vec{G}}
\newcommand{\mDel}{\vec{\Delta}}
\newcommand{\mPhi}{\vec{\Phi}}
\newcommand{\mGam}{\vec{\Gamma}}
\newcommand{\mN}{\vec{N}}
\newcommand{\mLam}{\vec{\Lambda}}

\newcommand{\vx}{\vec{x}}
\newcommand{\veta}{\vec{\eta}}
\newcommand{\vs}{\vec{s}}
\newcommand{\vp}{\vec{p}}
\newcommand{\vc}{\vec{c}}
\newcommand{\ve}{\vec{e}}

\newcommand{\vmu}{\vec{\mu}}
\newcommand{\vell}{\vec{\ell}}

\newcommand{\va}{\vec{a}}
\newcommand{\vz}{\vec{z}}
\newcommand{\vzt}{\overset{\sim}{\vec{z}}}
\newcommand{\zt}{\overset{\sim}{z}}
\newcommand{\vu}{\vec{u}}
\newcommand{\vv}{\vec{v}}
\newcommand{\vy}{\vec{y}}
\newcommand{\vzero}{\vec{0}}
\newcommand{\vone}{\vec{1}}
\newcommand{\vpsi}{\vec{\psi}}
\newcommand{\vphi}{\vec{\phi}}
\newcommand{\vxi}{\vec{\xi}}
\newcommand{\vrho}{\vec{\rho}}
\newcommand{\vlam}{\vec{\lambda}}
\newcommand{\vomega}{\vec{\omega}}

\newcommand{\Exp}{\mathsf{E}}

\newcommand{\Tr}{\mathsf{Tr}}

\newcommand{\cM}{\mathcal{M}}
\newcommand{\cN}{\mathcal{N}}

\newcommand{\cF}{\mathcal{F}}

\newcommand{\sO}{\mathcal{O}}

\newcommand{\bR}{\mathbb{R}}

\newcommand{\bZ}{\mathbb{Z}}

\newcommand{\ignore}[1]{{}}
\newcommand{\lse}{\text{lse}}
\newcommand{\gt}{\overset{\sim}{g}}

\newtheorem{thm}{Theorem}

\newtheorem{cor}{Corollary}

\newtheorem{pro}{Proposition}

\newcommand\blfootnote[1]{%
  \begingroup
  \renewcommand\thefootnote{}\footnote{#1}%
  \addtocounter{footnote}{-1}%
  \endgroup
}

\begin{document}

\title{Multimodal Data Fusion in High-Dimensional Heterogeneous Datasets via Generative Models}

\author{Yasin~Yilmaz\IEEEauthorrefmark{1},\;\; Mehmet Aktukmak\IEEEauthorrefmark{7},\;\;
        and \, Alfred O. Hero\IEEEauthorrefmark{7}}

\maketitle

\begin{abstract}
The commonly used latent space embedding techniques, such as Principal Component Analysis, Factor Analysis, and manifold learning techniques, are typically used for learning effective representations of homogeneous data. However, they do not readily extend to heterogeneous data that are a combination of numerical and categorical variables, e.g., arising from linked GPS and text data. In this paper, we are interested in learning probabilistic generative models from high-dimensional heterogeneous data in an unsupervised fashion. The learned generative model provides latent unified representations that capture the factors common to the multiple dimensions of the data, and thus enable fusing multimodal data for various machine learning tasks. Following a Bayesian approach, we propose a general framework that combines disparate data types through the natural parameterization of the exponential family of distributions. To scale the model inference to millions of instances with thousands of features, we use the Laplace-Bernstein approximation for posterior computations involving nonlinear link functions. The proposed algorithm is presented in detail for the commonly encountered heterogeneous datasets with real-valued (Gaussian) and categorical (multinomial) features. Experiments on two high-dimensional and heterogeneous datasets (NYC Taxi and MovieLens-10M) demonstrate the scalability and competitive performance of the proposed algorithm on different machine learning tasks such as anomaly detection, data imputation, and recommender systems. 
\end{abstract}


\blfootnote{\IEEEauthorrefmark{1}Electrical Engineering Department, University of South Florida, Tampa, FL 33620.\\
\IEEEauthorrefmark{7}Electrical Engineering and Computer Science Department, University of Michigan, Ann Arbor, MI 48109. The research in this paper was partially supported by a grant from the US Army Research Office under contract W911NF-15-1-0479 and a grant from the US National Science Foundation under award 2040572.}

\section{Introduction}
\label{sec:intro}

Finding lower-dimensional latent space representations of high-dimensional datasets is an important unsupervised learning problem for several objectives such as dimensionality reduction, visualization, exploratory data analysis, and data fusion. 
PCA is the most commonly used latent space embedding technique. It finds a succinct representation of the data points in terms of a smaller number of low-dimensional features obtained by linearly mixing the original features in such a way as to maximize variance. Such linear mixing coefficients are given by the eigenvectors of the covariance matrix of the feature variables that correspond to the $K$ largest eigenvalues, where $K$ is the desired number of new features ($K\le P$). For heterogeneous data, e.g., consisting of a numerical and a non-numerical variable, the standard sample covariance, called the Pearson covariance, is not directly applicable. For ordinal data Pearson introduced a modified covariance, called the polychoric correlation, that estimates the association between several ordinal variables by modeling them as quantized bivariate Gaussian random variables \cite{Pearson1900}. 
On the other hand, for mixed continuous and ordinal data, the polyserial correlation estimates association between the variables where, again, the ordinal data is modeled as quantized Gaussian \cite{Olsson82}.
PCA based on polyserial and polychoric correlations is often used for dimensionality reduction in heterogeneous datasets \cite{Kolenikov04,Kolenikov09}. However, both polyserial and polychoric correlations are designed for categorical variables that are ordinal, i.e., their values are linearly ordered \cite{Olsson82}. 
Following a probabilistic approach PCA can also be generalized to exponential family for homogeneous datasets \cite{Collins02,Mohamed09}.

Factor analysis (FA) is another well known latent space embedding technique, which includes PCA as a special case \cite{Bishop06}. The introduction of factor analysis is often attributed to Charles Spearman's work in 1904 \cite{Spearman04}, yet its roots can be traced to the earlier works of Francis Galton \cite{Galton1869}. Factor analysis is a latent variable model that decomposes a data matrix into low dimensional explanatory variables, called factors. Similar to PCA, factor analysis does not readily extend to heterogeneous data \cite[Ch. 5]{FA100}. In \cite{Khan10}, a mixture of factor analyzers is presented for heterogeneous data consisting of continuous and categorical variables. Similar to classical factor analysis, in \cite{Khan10}, instances (rows of the data matrix) are modeled independently with latent factor loading coefficients, whereas the features (columns of the data matrix) of an instance are linear combinations of factor loadings where the weights are called factor scores. 
Probabilistic Canonical Correlation Analysis (PCCA) \cite{Bach05} similarly models a pair of Gaussian feature vectors using a weighted sum of latent factors, called canonical components. Independent Component Analysis (ICA) and its extension Independent Vector Analysis (IVA), which is also a generalization of Canonical Correlation Analysis (CCA), are also used for multimodal data fusion \cite{Lahat15,Adali15}. 
Regarding multimodal data, similar to factor analysis and PCA, manifold learning methods, such as Laplacian eigenmaps \cite{Belkin02} and Isomap \cite{Tenenbaum00}, also require homogeneity among data points. Typically, such manifold learning methods require computing an affinity matrix, but it is not clear how to define a unified similarity or distance metric for disparate features (e.g., numerical and categorical). 
There is also a large literature on multi-view learning (e.g., \cite{Sun13,Wang13}) which aims at performing specific machine learning tasks using heterogeneous datasets. 
Some of these methods, such as parallel ICA, e.g., \cite{Liu09,Pearlson15}, have been applied to categorical data. However, in a heterogeneous data setting, these existing techniques \cite{{Bach05},Lahat15,Adali15,Belkin02,Tenenbaum00,Sun13,Wang13} treat the categorical data in the same way as continuous-valued data, which is a strictly suboptimal approach.

There are also methods which address specific heterogeneous data applications using latent factor models, e.g., \cite{ray2014bayesian,Salazar2014}. In \cite{Yang2012}, a generic method based on generalized linear models is proposed to build graphical models from the exponential family of distributions. However, a joint analysis of the aggregated exponential family is not discussed in \cite{Yang2012}. While the mixtures of factor analyzers methods proposed in \cite{Ghahramani1996,Ghahramani2000} jointly analyze heterogeneous Gaussian data, they do not allow for heterogeneous data types, such as numerical and categorical, as the proposed mixture is based on classical factor analyzers. 

This paper develops a general approach to joint factor analysis for heterogeneous data, called multimodal factor analysis (MMFA). MMFA, originally introduced in \cite{MLSP}, is a Bayesian approach that models different types of data using latent factor loadings specific to each data type and latent factor scores that are common to the data types. 
It was applied to event detection in Twitter \cite{MMFA-Twitter} in order to fuse categorical and spherical data that are modeled by multinomial and von Mises-Fisher distributions, respectively. 

In this paper, we present MMFA as a comprehensive unsupervised learning tool for learning generative models in high-dimensional and heterogeneous datasets explainable by exponential family of distributions. Specifically, our contributions can be summarized as follows.
\bi
\item As opposed to the preliminary work \cite{MLSP,MMFA-Twitter}, the proposed generalized MMFA model provides a tractable unified framework for jointly analyzing heterogeneous features from the exponential family of distributions through linking 
their natural parameters with a common factor score vector. 
\item Motivated by the Bernstein-von Mises theorem \cite{Vaart00} MMFA finds Gaussian approximations to the posterior distribution of latent factor loadings for the data types whose natural parameters require nonlinear link functions, e.g., multinomial distribution (a.k.a. Laplace-Bernstein approximation). 
\item MMFA easily scales to high-dimensional datasets with many heterogeneous features and instances, as demonstrated by the experimental results, thanks to 
the Laplace-Bernstein approximations.
\ei

The problem formulation is given in Section \ref{sec:prob}, and the proposed generalized MMFA model is introduced in Section \ref{sec:model}. 
In Section \ref{sec:learn}, a variational learning algorithm for the proposed model is presented. In Section \ref{sec:gaus_mult}, the MMFA algorithm is illustrated for a heterogeneous dataset consisting of Gaussian and multinomial components, and an analysis of computational complexity and mean-squared-error (MSE) performance is presented.
We also demonstrate the usage of MMFA in unsupervised learning problems using real datasets (Section \ref{sec:exp}). Finally, the paper is concluded in Section \ref{sec:conc}.


\section{Problem Formulation}
\label{sec:prob}

Consider a heterogenous random data structure composed of $M$ different data types, called modalities, from the same source. Observed are $P$ realizations of this data structure, called instances. Assume that the data from each modality can be modeled with a probability distribution from the {\it exponential dispersion family} (e.g., Gaussian, Poisson, multinomial), which is a generalization of the exponential family \cite{Jorgensen87}. For each modality $m$ and each instance $i$, if the modality corresponds to  a continuous random variable of dimension $D_m$ then it can be represented as a data vector $\vx_{im}=[x_{im}^1 \ldots x_{im}^{D_m}]^T$ with probability density function (pdf) given by
\be
\begin{split}
\label{eq:exp_cont}
	f_m(\vx_{im} | \veta_m, \tau_m) &= h_m(\vx_{im},\tau_m) \\& \exp\left\{ \tau_m \left[\veta_m^T ~\vs_m(\vx_{im})  - a_m(\veta_m) \right]  \right \};
\end{split}
\ee
if $\vx_{im}$ is discrete-valued, its probability mass function (pmf) is given by
\be
\begin{split}
\label{eq:exp_disc}
	f_m(\vx_{im} | \veta_m, \tau_m) &= h_m(\vx_{im},\tau_m) \\& \exp\left\{ \veta_m^T ~\vs_m(\vx_{im}) - \tau_m a_m(\veta_m)  \right\}.
\end{split}
\ee

In \eqref{eq:exp_cont} and \eqref{eq:exp_disc}, for modality $m$, $\veta_m$ is a vector of natural parameters, $\vs_m(\vx_{im})$ is a vector of sufficient statistics, $a_m(\veta_m)$ is the log-partition (i.e., log-normalization) function, $\tau_m\ge0$ is the dispersion parameter, $(\cdot)^T$ denotes the transpose, and $i = 1,\ldots,P, ~m = 1,\ldots,M$.
Note that the dimension $D_m$ of each modality can be different, and $D=\sum_{m=1}^M D_m$ gives the total number of dimensions, i.e., features, in the dataset. 

Table \ref{tab:exp} gives examples of some popular probability distributions from the exponential family in terms of the representations in \eqref{eq:exp_cont} or \eqref{eq:exp_disc}. 
The first three and the last three rows of Table \ref{tab:exp} refer to continuous cases and discrete cases, i.e., \eqref{eq:exp_cont} and \eqref{eq:exp_disc}, respectively.


\begin{table*}[t]
\begin{threeparttable}[b]
\centering
\caption{Examples of exponential dispersion family.}
\label{tab:exp}
\def\arraystretch{1.5}
\begin{tabular}{| p{0.15\linewidth} | p{0.30\linewidth} | p{0.11\linewidth} | p{0.04\linewidth} | p{0.04\linewidth} | p{0.11\linewidth} | p{0.09\linewidth} |}
\hline
& pdf / pmf & $\veta$ & $\tau$ & $\vs(\vx)$ & $a(\veta)$ & $h(\vx,\tau)$  \\ 
\hline
\centering Gaussian $(\mu,\sigma^2)$ & $\frac{\exp(-(x-\mu)^2/2\sigma^2)}{\sqrt{2\pi\sigma^2}}$ & $\mu$ & $1/\sigma^2$ & $x$ & $\eta^2/2$ & $\frac{\exp(-x^2 \tau/2)}{\sqrt{2\pi/\tau}}$ \\
\hline
\centering Exponential $(\lambda)$ & $\lambda \exp(-\lambda x)$ & $-\lambda$ & $1$ & $x$ & $-\log(-\eta)$ & $1$ \\
\hline
\centering von Mises-Fisher $(\vmu^{d \times 1}, \kappa)$\tnote{1} & $C_d(\kappa)\exp(\kappa \vmu^T\vx)$ & $\vmu$ & $\kappa$ & $\vx$ & $0$ & $C_d(\tau)$ \\
\hline
\centering Poisson $(\lambda)$ & $\lambda^x \exp(-\lambda)/x!$ & $\log\lambda$ & $1$ & $x$ & $\exp(\eta)$ & $1/x!$ \\
\hline
\centering Binomial $(n,p)$ & $\frac{n!}{x!(n-x)!}p^x(1-p)^{n-x}$ & $\log\frac{p}{1-p}$ & $n$ & $x$ & $\log(1+e^\eta)$ & $\frac{\tau!}{x!(\tau-x)!}$ \\
\hline
\centering Multinomial $(n,\vp^{d\times1})$ & $\frac{n!}{x_1! \cdots x_d!}p_1^{x_1} \cdots p_d^{x_d}$ & $\left[\begin{array}{c}\log\frac{p_1}{p_d}\\ \vdots \\ \log\frac{p_{d-1}}{p_d} \\ 0 \end{array}\right]$ & $n$ & $\vx$ & \vspace{-8mm}$$\log\Big(\sum_{j=1}^{d}e^{\eta_j}\Big)$$ & $\frac{\tau!}{x_1! \cdots x_d!}$ \\
\hline
\end{tabular}
\begin{tablenotes}
            \item [1] The von Mises-Fisher distribution is an extension of the Gaussian distribution to spherical data. For the $d$-dimensional von Mises-Fisher distribution, $C_d(\kappa)=\frac{\kappa^{d/2-1}}{(2\pi)^{d/2} I_{d/2-1}(\kappa)}$, where $I_{d/2-1}$ is the modified Bessel function of the first kind at order $d/2-1$, e.g., $C_3(\kappa)=\frac{\kappa}{2\pi(e^\kappa-e^{-\kappa})}$.
\end{tablenotes}
\end{threeparttable}
\end{table*}


\section{Proposed Generative Latent Variable Model}
\label{sec:model}

\begin{figure}
\centering \includegraphics[width=1.0\linewidth]{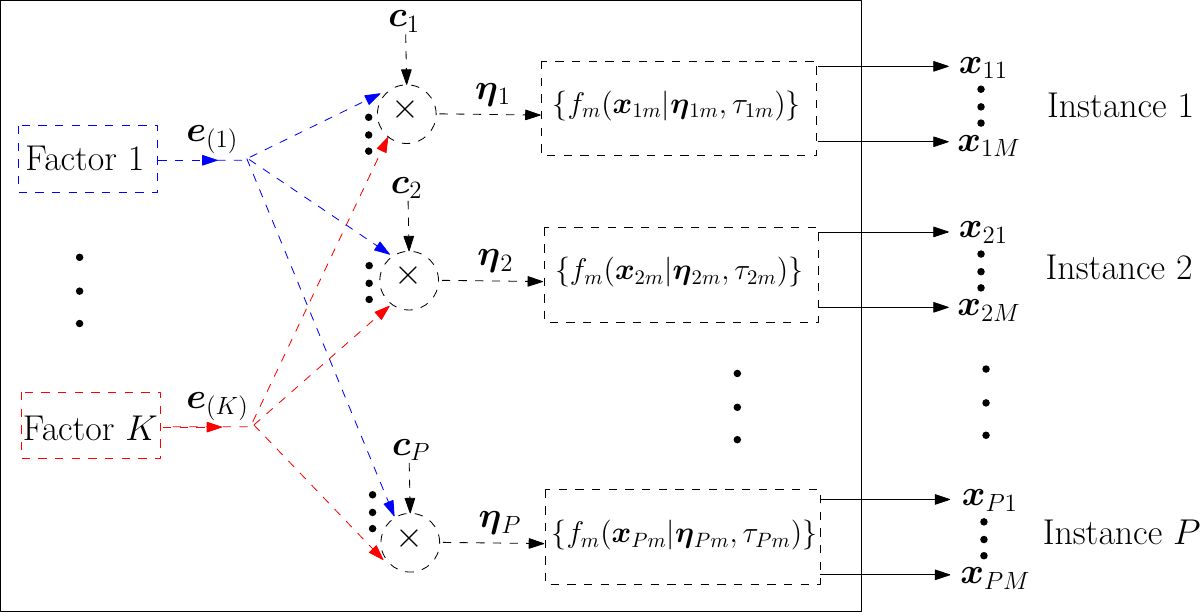} \\ \vspace{3mm}
\caption{Operational diagram of the considered system model. The outer rectangle represents the entire system; and the dashed part represents the proposed generative factor-based model, characterized by the latent vectors $\{\ve_{(1)},\ldots,\ve_{(K)}\}$, $\{\vc_{1},\ldots,\vc_{P}\}$, and the exponential family distributions $\{f_{1},\ldots,f_{M}\}$.} 
\label{fig:oper}
\end{figure}

Given a data structure of $M$ modalities, $D$ dimensions, and $P$ instances, the MMFA model summarizes the data with a small number of $K$ latent factors where $K\ll \min\{D,P\}$. This generative model is illustrated in Fig. \ref{fig:oper} where the latent factors $\{e_{(k)}\}$ and the $P$ instances are shown. Fig. \ref{fig:oper} is a Markov graph in the sense that, conditioned on the $e_{(k)}$'s, the instances are independent and modalities are distributed according to different exponential family distributions of the form \eqref{eq:exp_cont} and \eqref{eq:exp_disc}.

Each instance $i$ is assumed to follow an exponential model of the form \eqref{eq:exp_cont} or \eqref{eq:exp_disc} with a  natural parameter vector $\veta_i = [\veta_{i1}^T \cdots \veta_{im}^T]^T\in \bR^{\overset{\sim}{M}}$ composed of an instance-wide  latent matrix $\mE$ and an instance-specific score vector $\vc_i \in \bR^K$: 
\be
	\veta_i = \mE^T \vc_i, \nn
\ee	
where $\mE = [\ve_1 \cdots \ve_{\overset{\sim}{M}}]$, $\ve_m \in \bR^K$ are latent vectors, $\overset{\sim}{M}$ is the total number of natural parameters used to model the multimodal data \footnote[2]{If every modality in the observation model has a single natural parameter, then $\overset{\sim}{M}=M$.}. 
The latent vectors $\{\ve_{(1)},\ldots,\ve_{(K)}\}$ in Fig. \ref{fig:oper} are defined as the rows of the matrix $\mE$. 
In analogy with other factor analysis methods, the latent vectors $\{\ve_m\}$ are called factor loading vectors and $\{\vc_i\}$ are called factor score vectors. 
In the proposed model, each instance $i$ is characterized by the score vector $\vc_i$. The same $\vc_i$ is used to linearly model each natural parameter $\eta_{im}$ associated with instance $i$
for all of the $M$ data modalities, e.g., \eqref{eq:gaus}-\eqref{eq:binom}. 

For example, if the first three modalities are Gaussian, Poisson, and binomial, respectively, then from Table \ref{tab:exp},  the natural parameter $\veta_i$ is composed of three elements:
\begin{align}
\label{eq:gaus}
	\mu_i &= \ve_1^T \vc_i, \\
\label{eq:pois}
	\log \lambda_i &= \ve_2^T \vc_i, \\
\label{eq:binom}
	\log \frac{p_i}{1-p_i} &= \ve_3^T \vc_i.
\end{align}

The Gaussian case in \eqref{eq:gaus} corresponds to the classical factor analysis model. 
Factor analysis, in its original form, is used to model the mean of continuous data through a linear combination of continuous latent variables \cite{FA100}, as in the Gaussian case given by \eqref{eq:gaus}. 
The classical factor analysis model does not provide a good fit for discrete data (e.g., Poisson data) or categorical data (e.g., binomial data) \cite{Bartho11}. 
There are several latent variable models that extend factor analysis to non-Gaussian data, such as the general linear latent variable model \cite{Bartho11}, Poisson factor analysis \cite{Zhou12}, and latent Dirichlet allocation \cite{lda}.
Different than those works, here we provide a joint model to deal with different data modalities together. 
In our proposed model, instead of the mean -- e.g., $\lambda$ for Poisson and $p$ for binomial (see Table \ref{tab:exp}) -- we linearly model the natural parameter -- e.g., $\log\lambda$ for Poisson \eqref{eq:pois} and $\log\frac{p}{1-p}$ for binomial \eqref{eq:binom} -- which provides a general framework for the exponential family of distributions. 
This mapping is similar to the general linear latent variable model \cite{Bartho11} and the generalized mixture of factor analyzers model \cite{Khan10}, albeit with some important differences. 

Firstly, we present a generic model for the joint analysis of exponential family where either the factor loadings $\{\ve_m\}$ or the factor scores $\{\vc_i\}$ can be modeled as latent variables, whereas they are strictly modeled as parameters and latent variables, respectively, in the existing factor analysis models including \cite{Bartho11} and \cite{Khan10}. The proposed model can bring about significant tractability for high-dimensional heterogeneous datasets since computing the posterior of $\vc_i$ involves all data modalities, whereas computing the posterior of $\ve_m$ only requires data from modality $m$. 
We also provide a general scalable framework based on the Laplace-Bernstein approximation for fitting the proposed model to high-dimensional and heterogeneous datasets.



The proposed model is indeed a nonparametric model. Although a parametric model is used for each data vector $\vx_{im}$ of instance $i$ and modality $m$ (see \eqref{eq:exp_cont}, \eqref{eq:exp_disc}), we have a nonparametric model for the entire dataset $\{\vx_{im}\}$ representing the collection of instances. This is because the number of parameters $\{\vc_i\}$ linearly increases with the number of instances, and thus is asymptotically infinite. 

\ignore{
\section{Connection with CNN, RBM and Deep Learning}
\label{sec:RBM}

\begin{figure}
\centering \includegraphics[width=.75\linewidth]{GCNN.pdf}
\caption{Convolutional neural network representation of the proposed generative latent variable model, given by \eqref{eq:model_vec}. The standard parameters can be obtained by applying a nonlinear transformation layer to the natural parameters $\{\eta_{im}\}$, e.g., $\lambda_i = e^{\eta_{i2}}$ for Poisson in \eqref{eq:pois}, and $p_i = \frac{e^{\eta_{i3}}}{1+e^{\eta_{i3}}}$ for binomial in \eqref{eq:binom}.}
\label{fig:gcnn}
\end{figure}

The structure of the proposed generative latent variable model can be showb as a convolutional neural network (CNN) (see Fig. \ref{fig:gcnn}). A hypothetical image that consists of the coefficient vectors $\{\vc_i\}$ of the observed instances is passed through a number of filters each of which represents a different modality. The outcome of each filter $m$ is the natural parameters $\{\eta_{im}\}_i$ of the instances in this modality. Finally, the data vector $\vx_{im}$ is a realization from the parameterization $\eta_{im}$. Note that CNN is typically a feed-forward network in which the known entity is the input and the unknown entities are the filter weights and the output, whereas MMFA is a generative model that knows the output and learns the input and the coefficients. Keeping in mind this fundamental operational difference we only show here the structural analogy between MMFA and CNN.

\begin{figure}
\centering \includegraphics[width=.7\linewidth]{OFA.pdf}
\caption{MMFA as a directed counterpart of RBM.}
\label{fig:ofa}
\end{figure}

Restricted Boltzmann Machine (RBM) is an undirected generative model, i.e., Markov random field (MRF). MMFA can be seen as a directed counterpart of RBM. As shown in Fig. \ref{fig:ofa}, factors are in the hidden layer and instances are in the visible layer. Factors have a number of latent variables which are connected to the instances' observed variables through coefficient vectors $\{\vc_i\}$. There is a single coefficient vector for each instance. The number of variables, $\overset{\sim}{M}$, is the same for factors and instances. The output of instance variables are possibly passed through a nonlinear link function before the random observations are produced.  

Alternatively, following the same idea presented in this paper RBMs can be used for multimodal data fusion, which we plan to investigate as future work. 
Adding nonlinear link (i.e., activation) functions also to the latent variables $\{e_{im}\}$ in hidden layer we can obtain different types of RBM. For instance, using logistic sigmoid function we get binary (Bernoulli) latent variables instead of Gaussian \cite[p. 989]{Murphy12}. By adding more hidden layers (i.e., layers of factors) we can obtain ``deep" versions of the proposed generative model similarly to the Deep Boltzmann Machines or Deep Belief Networks. We reserve the deep learning aspect of the proposed multimodal data fusion technique to a future work. 

RBM is typically used to learn a representation of an instance from multiple observations of a single feature, such as the instensity of image pixels. For each observation a coefficient vector is learned, and a factor score vector is learned to encode the feature, which is then used as the new representation of the instance. In this paper, we present a way to learn a representation from multiple observations of multiple features by using the same coefficient vector for each feature. It is not clear how to do this using a standard RBM for multiple features. 
}

\section{Learning the Model Parameters}
\label{sec:learn}

We use the expectation-maximization (EM) approach to compute, in an iterative manner, the maximum likelihood estimates of the model parameters $\theta$, which includes the score vectors $\{\vc_i\}$, the hyperparameters of the prior distribution for $\{\ve_m\}$, and the dispersion parameters $\{\tau_m\}$. 
The number of factors $K$ can be selected in different ways, including Bayesian information criterion (BIC), Monte Carlo integration over a uniform on $K$, birth-death models, or evaluation of a knee in the scree plot of goodness of fit, e.g., as measured by the negative log-likelihood. 

\subsection{The EM Algorithm}
\label{sec:em}

In the original EM algorithm \cite{Dempster77} for modality $m$, at each iteration $n$, in the expectation step (E-step), the expectation of the complete-data log-likelihood is computed
with respect to the posterior distribution 
$	g_{m,n-1}\big(\ve_m | \{\vx_{im}\}_i, \theta_{m,n-1}\big)$ 
of each latent vector $\ve_m$, 
i.e., $Q_n(\theta_m) = \Exp_{g_{m,n-1}}\big[ \log f_m\big(\{\vx_{im}\}_i, \ve_m | \theta_m\big) \big]$. Then, in the maximization step (M-step), the parameters $\theta_{m,n}$ are estimated by maximizing $Q_n(\theta_m)$ 
over the parameter space $\Theta_m$, i.e.,
$	\theta_{m,n} = \arg\max_{\theta_m \in \Theta_m} Q_n(\theta_m)$. 


\ignore{
In the E-step, given by \eqref{eq:E-step}, the terms that include the unknown vector $\ve_m$ in the complete-data log-likelihood are replaced with their posterior expected values. In the M-step, we maximize the following lower bound to the log-likelihood 
$L(\theta_m) = \log f_m(\{\vx_{im}\}_i | \theta_m)$,
\begin{align}
\label{eq:Jensen}
	L(\theta_m)  &= \log \int_{\ve_m} f_m\big(\{\vx_{im}\}_i, \ve_m | \theta_m\big) ~ \text{d}\ve_m \nn\\
	&= \log \Exp_{g_{m,n-1}}\left[ \frac{f_m\big(\{\vx_{im}\}_i, \ve_m | \theta_m\big)}{g_{m,n-1}(\ve_m | \{\vx_{im}\}_i, \theta_{m,n-1})} \right] \nn\\
	&\ge \Exp_{g_{m,n-1}}\left[ \log \frac{f_m\big(\{\vx_{im}\}_i, \ve_m | \theta_m\big)}{g_{m,n-1}(\ve_m | \{\vx_{im}\}_i, \theta_{m,n-1})} \right] \\
	&= Q_n(\theta_m) - \Exp_{g_{m,n-1}}\left[ \log g_{m,n-1}(\ve_m | \{\vx_{im}\}_i, \theta_{m,n-1}) \right],
\end{align}
which holds due to the Jensen's inequality. Note from \eqref{eq:post_g} that we have equality in \eqref{eq:Jensen} for $\theta_m = \theta_{m,n-1}$. Therefore, in the M-step, when we maximize, over $\theta_m$, the lower bound in \eqref{eq:Jensen}, or equivalently $Q_n(\theta_m)$ in \eqref{eq:M-step}, we increase the log-likelihood, i.e., 
\begin{align}
\label{eq:converge}
	L(\theta_{m,n}) &\ge Q_n(\theta_{m,n}) - \Exp_{g_{m,n-1}}\left[ \log g_{m,n-1}(\ve_m | \{\vx_{im}\}_i, \theta_{m,n-1}) \right] \nn\\
	&\ge Q_n(\theta_{m,n-1}) - \Exp_{g_{m,n-1}}\left[ \log g_{m,n-1}(\ve_m | \{\vx_{im}\}_i, \theta_{m,n-1}) \right] 
	= L(\theta_{m,n-1}), ~\forall n.
\end{align}
The EM algorithm is guaranteed to converge to a stationary point
\be
	\lim_{n \to \infty} L(\theta_{m,n}) = L^*,
\ee
which is either a global or local maximum, or a saddle point \cite{Wu83}. 
}

With the EM approach, the biggest challenge is to compute the posterior distribution 
in the E-step at each iteration $n$ since the link function relating the natural parameters to latent vectors is in general nonlinear, and the posterior is multivariate. 
Markov chain Monte Carlo (MCMC) methods, such as the Gibbs sampler, can be used to sample from the posterior $g_{m,n-1}(\ve_m | \{\vx_{im}\}_i, \theta_{m,n-1})$. We might then use these samples to estimate $Q_n(\theta_m)$ 
for a given $\theta_m$, and search over the parameter space to find the $\theta_m$ that maximizes $Q_n(\theta_m)$. 
Although the MCMC approach can enable the use of exact EM algorithm for inferring the model parameters, high-dimensional parameter space with thousands of features (large $D$) and millions of instances (large $P$) could be problematic for the convergence rate and computational complexity. 
We propose a variational EM approach to address this problem.

\subsection{The Proposed Variational EM Algorithm}
\label{sec:var_em}


In the variational EM approach, a tractable probability distribution $\overset{\sim}{g}_{m,n-1}(\ve_m | \{\vx_{im}\}_i, \theta_{m,n-1})$ is used to approximate the posterior. The objective is to select, from a tractable family of distributions, the distribution which is closest to the actual posterior in the KL-divergence sense, i.e., $\overset{\sim}{g}_{m,n-1}=\arg\min_{q} \text{KL}(q||g_{m,n-1})$. The design challenge here is to determine the tractable family of distributions such that $\overset{\sim}{g}_{m,n-1}$ will be close to $g_{m,n-1}$. To this end we utilize the Bernstein-von Mises theorem, which states that the posterior is asymptotically, as the number of instances $P$ increases, well approximated by a Gaussian distribution when the likelihood model is from exponential family and the prior is Lipschitz continuous, e.g., Gaussian, von Mises-Fisher, etc. \cite{Vaart00,Belloni14}. In the problem of interest with exponential family models, Gaussian priors, and large number of instance-feature interactions (e.g., $P \times D$ is on the order of millions), the Bernstein-von Mises theorem provides theoretical motivation for using Gaussian approximation to the posterior. Note that $P \times D$ gives the number of observations for finding the posterior of the $K$-dimensional latent vectors. Hence, following a variational EM approach, we approximate $g_{m,n-1}$ with a Gaussian $\overset{\sim}{g}_{m,n-1}$.

The Laplace technique approximates the posterior with the Gaussian $\cN(\vell,-\mH(\vell)^{-1})$, where $\vell$ is the mode of the posterior, 
and $\mH(\vell)$ is the Hessian matrix (i.e., second-order derivative of $g_{m,n-1}$ with respect to $\ve_m$) evaluated at the mode $\vell$. 
The posterior mode typically does not have a closed form due to the nonlinear link function, and thus calls for iterative computation through a numerical optimization technique. The computation of the posterior mode and the Hessian matrix at each EM iteration, with a high-dimensional dataset, may be prohibitive. 
Similarly, higher-order variational inference techniques such as expectation propagation (EP) may incur significant computational complexity in high-dimensional datasets (large $P$, large $D$). 
Through moment matching EP will need to iteratively compute the mean vector and covariance matrix of the actual posterior $g_{m,n-1}$. 

Hence, for scalability to large datasets, we resort to variational -- in particular, quadratic -- lower bounds, which significantly reduces the computational complexity compared to the Laplace approximation and EP by fixing the covariance matrix. Moreover, as noted in \cite[p. 498]{Bishop06}, the variational lower bound method has additional flexibility, which leads to improved accuracy, compared to the Laplace method.
This motivates us to approximate the log-partition function $a(\eta)$, which is the problematic term in the complete-data log-likelihood $\log f_m\big(\{\vx_{im}\}_i, \ve_m | \theta_m\big)$ with a quadratic term to obtain the Gaussian approximation $\overset{\sim}{g}_{m,n-1}$. 

For instance, we approximate $\log(-\eta)$ and $e^\eta$ in the exponential and Poisson likelihoods (see Table \ref{tab:exp}) using the second-order Taylor series expansion around $1$ and $0$, respectively. Specifically, we obtain an evidence lower bound (ELBO) by using $\min\{-1,-x_i^2\}$ and $\max\{1,x_i\}$ for the second-order derivative term, where $x_i$ approximates $1/\lambda$ and $\lambda$ in the exponential and Poisson case, respectively. In the next section, we will explain this procedure in detail for the multinomial likelihood, which is commonly used in the real-world datasets for categorical features. A Gaussian prior $\pi(\ve_m)$ is assumed for each modality to facilitate the Gaussian approximation for the posterior. Note that no approximation to the posterior is needed for the Gaussian likelihood since the posterior is already Gaussian with a conjugate prior. For the von Mises-Fisher distribution, which is an extension of Gaussian distribution to spherical data, we use a von Mises-Fisher prior, which is conjugate to the likelihood. However, in this case, an approximation is still needed due to the constraint that the mean vector is a unit-length vector (see \cite{MMFA-Twitter} for details).

\begin{figure}
\centering \includegraphics[width=0.9\linewidth]{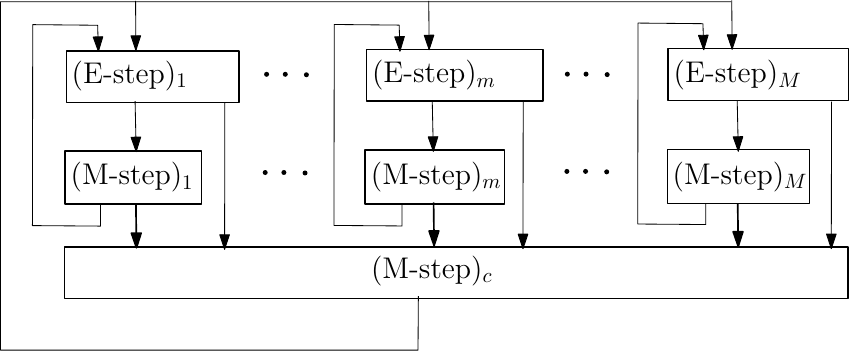}
\caption{Structure of the proposed variational EM algorithm. In each iteration, E-M steps of all modalities run in parallel. They are synchronized by a final M step for the score vector of each instance.}
\label{fig:em}
\end{figure}

In the proposed variational EM algorithm, for all modalities EM steps are run in parallel, which is followed by the M-step for the score vectors $\{\vc_i\}$, as shown in Fig. \ref{fig:em}. Since $\vc_i$ is common to all modalities of instance $i$, it is updated by receiving related information from all EM steps for different modalities. Each $\vc_i$ can be updated in parallel.




\section{Example: Gaussian and Multinomial}
\label{sec:gaus_mult}

To illustrate the proposed model and variational EM algorithm, in this section, we consider a bimodal dataset $\mX=[\mY \mZ]$ from the same source consisting of a real-valued matrix $\mY \in \bR^{P \times D_1}$ and a categorical data matrix $\mZ \in \bZ_+^{P \times D_2}$ with $P$, $D_1$, and $D_2$ denoting the number of instances, number of real-valued features, and the number of categories, respectively. $\bZ_+$ denotes the set of nonnegative integers. Assume the entries $y_{ij}$ of $\mY$ and the rows $\vz_{(i)}$ of $\mZ$ are well modeled using the Gaussian $\cN(\mu_{ij},\sigma_{ij}^2)$ and the multinomial $\cM(N_i,\vp_i)$ models (where $N_i$ is the number of experiments and $\vp_i \in [0,1]^{D_2}$ is the probability vector), respectively. 

As in \eqref{eq:gaus} and \eqref{eq:binom}, assuming $K$ generative factors that are characterized by the latent vectors 
\begin{align}
	\{\ve_{(k)}&: \ve_{(k)} \in \bR^{D_1+D_2-1}, k=1,\ldots,K\}, \nn\\
	\mE = [\mU \mV] &= [\vu_1 \cdots \vu_{D_1} \vv_1 \cdots \vv_{D_2-1}] = [\ve_{(1)} \cdots \ve_{(K)}]^T, \nn	
\end{align}
we linearly model the natural parameters of the Gaussian and the multinomial distributions, i.e.,
\begin{align}
	\mu_{ij} &= \vu_j^T \vc_i, ~ j=1,\ldots,D_1 \nn\\
	\log \frac{p_{ij}}{p_{iD_2}} &= \vv_j^T \vc_i, ~ j=1,\ldots,D_2-1. \nn
\end{align}
The coefficient vector $\vc_i$ represents each instance $i$ in terms of the $K$ latent factors. In the multinomial distribution, the last category is selected as pivot since one of the probabilities is fully determined by the other $D_2-1$ probabilities, i.e., the degree of freedom is one less than the number of categories (cf. Table \ref{tab:exp}). Note that multinomial distribution covers as a special case categorical distribution ($N_i=1, \forall i$), binomial distribution ($D_2=2$), and Bernoulli distribution ($D_2=2, N_i=1, \forall i$). The two modalities considered in this section are the most common data types found in real-world datasets \cite{UCI}.

\subsection{Model Parameters}

\subsubsection{Gaussian Parameters}
\label{sec:gaus}

For the Gaussian model, assuming the conjugate prior $\cN(\vzero_K,\mI_K)$ for $\vu_j, j=1,\ldots,D_1$, where $\vzero_K$ and $\mI_K$ denote the $K$-dimensional zero vector and identity matrix, we have the exact EM algorithm, i.e., at each iteration $n$
\begin{align}
	\text{E-step:} & ~ \text{compute the posterior} \nn\\ & g_{j,n-1}\big(\vu_j | \vy_j, \theta_{j,n-1}\big) = \frac{f_1\big(\vy_j, \vu_j | \theta_{j,n-1}, \vc_{i,n-1} \big)}{f_1\big(\vy_j | \theta_{j,n-1}, \vc_{i,n-1} \big)} \nn\\
	\text{M-step:} & ~ \text{estimate the parameters} \nn\\ & \theta_{j,n} = \arg\max_{\theta_j \in \Theta_j} \Exp_{g_{j,n-1}}\Big[ \log f_1\big(\vy_j, \vu_j | \theta_j, \vc_{i,n-1} \big) \Big], \nn
\end{align}
where $\theta_j = \{\sigma_{ij}^2\}_{i=1,\ldots,P}$. 
From the classical factor analysis \cite{FA100,Murphy12}, we have $\cN(\va_{j,n}, \mB_{j,n})$ as the posterior of $\vu_j$ at iteration $n$, where
\begin{align}
	\mB_{j,n} &= \left( \mC_{n-1} \mSig_{j,n-1}^{-1} \mC_{n-1}^T + \mI_K \right)^{-1}, \nn\\ \va_{j,n} &= \mB_{j,n} \mC_{n-1} \mSig_{j,n-1}^{-1} \vy_j, \label{eq:gaus-e}\\
	\mC_{n-1} &= \Big[ \vc_{1,n-1} \cdots \vc_{P,n-1} \Big], \nn\\ \mSig_{j,n-1} &= \text{diag}\Big( \sigma_{1j,n-1}^2 \cdots \sigma_{Pj,n-1}^2 \Big), \nn
\end{align}
and the parameter update
\begin{align}
\sigma_{ij,n}^2 &= \left( y_{ij}-\va_{j,n}^T\vc_{i,n-1} \right)^2 + \vc_{i,n-1}^T \mB_{j,n} \vc_{i,n-1}. \nn
\end{align}
Assuming an inverse-gamma distribution $InvGam(\alpha,\beta)$, which is the conjugate prior for the Gaussian variance, for $\sigma_{ij}^2$ it is straightforward to show that the update becomes
\begin{align}
\sigma_{ij,n}^2 &= \frac{\left( y_{ij}-\va_{j,n}^T\vc_{i,n-1} \right)^2 + \vc_{i,n-1}^T \mB_{j,n} \vc_{i,n-1} + 2/\beta}{2(\alpha+1)+1}. \label{eq:gaus-m}
\end{align}

Combining the inputs from all features (Gaussian and multinomial) the coefficient vector $\vc_i$ of instance $i$ is updated as
\begin{align}
\vc_{i,n} = &\arg\max_{\vc_i} -\frac{1}{2} \vc_i^T \left( \sum_{j=1}^{D_1} \frac{\mB_{j,n}+\va_{j,n}\va_{j,n}^T}{\sigma_{ij,n}^2} \right) \vc_i \nn\\ &+ \vc_i^T \sum_{j=1}^{D_1} \frac{y_{ij}\va_{j,n}}{\sigma_{ij,n}^2} + \zeta_n(\vc_i),
\label{eq:c-m-gaus}
\end{align}
where the input from the multinomial features $\zeta(\vc_i)$ will be derived next. 

\subsubsection{Multinomial Parameters}

In the multinomial likelihood, 
\begin{align}
f_2\big( \{\vz_{(i)}\} | \{\vv_j\}, &\vc_i, N_i \big) = \prod_{i=1}^P \frac{N_i!}{z_{i1}!\cdots z_{iD_2}!} \prod_{j=1}^{D_2} p_{ij}^{z_{ij}} \nn\\
&= \prod_{i=1}^P \frac{N_i!}{z_{i1}!\cdots z_{iD_2}!} \frac{\prod_{j=1}^{D_2-1} e^{\vv_j^T\vc_iz_{ij}} }{\left(1+\sum_{j'=1}^{D_2-1}e^{\vv_{j'}^T\vc_i} \right)^{N_i} }, \nn
\end{align}
the dependency of the sum-of-exponentials (sum-exp) term in the denominator on all latent vectors $\{\vv_j\}$ complicates the analysis considerably. First of all, we need to consider $\{\vv_j\}$ together and obtain the posterior distribution of the combined latent vector $\vv=[\vv_1^T \cdots \vv_{D_2-1}^T]^T$, which is $K(D_2-1)$-dimensional. More importantly, finding the exact posterior is not tractable due to the presence of $\{\vv_j\}$ in the sum-exp function. We rewrite the above likelihood expression in a more compact form 
\begin{align}
f_2\big( \{\vz_{(i)}\} | \{\vv_j\}, & \vc_i, N_i \big) \nn\\ & = \prod_{i=1}^P \frac{N_i!}{z_{i1}!\cdots z_{iD_2}!}  e^{\sum_{j=1}^{D_2-1} \vv_j^T\vc_iz_{ij} - N_i \lse(\veta_i)}, \nn\\
&= \prod_{i=1}^P \frac{N_i!}{z_{i1}!\cdots z_{iD_2}!}  e^{\vv^T\mC_i \vz_i - N_i \lse(\veta_i)}, \nn
\end{align}
where $\lse(\veta_i) = \log\left(1+\sum_{j'=1}^{D_2-1}e^{\vv_{j'}^T\vc_i} \right)$ is the log-sum-exp function, $\veta_i = [\vv_1^T\vc_i \cdots \vv_{D_2-1}^T\vc_i]^T$, $\mC_i = \mI_{D_2-1} \otimes \vc_i$ with $\otimes$ being the Kronecker product, and $\vz_i = [z_{i1} \cdots z_{iD_2-1}]^T$.

To obtain an approximate posterior $\gt_{2,n}(\vv | \{\vz_i, \vc_i, N_i\})$, as outlined in Section \ref{sec:var_em}, we derive a lower bound for the likelihood, and accordingly for the complete-data log-likelihood, through the second-order Taylor series expansion of $\lse(\veta_i)$ around a fixed point $\vpsi_i$,
\begin{align}
\lse(\veta_i) &= \lse(\vpsi_i) + (\veta_i-\vpsi_i)^T \nabla \lse(\vpsi_i) \nn\\ & \ \ \ + \frac{1}{2} (\veta_i-\vpsi_i)^T \nabla^2 \lse(\vpsi_i + \epsilon(\veta_i-\vpsi_i)) (\veta_i-\vpsi_i) \nn\\
&\le \lse(\vpsi_i) + (\veta_i-\vpsi_i)^T \nabla \lse(\vpsi_i) \nn\\ & \ \ \ +\frac{1}{2} (\veta_i-\vpsi_i)^T \mA (\veta_i-\vpsi_i),
\label{eq:lse}
\end{align}
where $\epsilon \in [0,1]$, and $\mA = \frac{1}{2}\left( \mI_{D_2-1}-\frac{\vone\vone^T}{D_2} \right)$ from \cite{Bohning92}. Note that we defined a new variable $\vpsi_i$ for each instance $i$. We will show how to update it in Proposition \ref{pro:mult}. The gradient is given by the probability vector induced by $\vpsi_i$, i.e., 
$\nabla \lse(\vpsi_i) = \frac{e^{\vpsi_i}}{1+\sum_{j=1}^{D_2-1} e^{\psi_{ij}}} = \vp_{\vpsi_i}.$ 
Replacing the lse function with this quadratic upper bound we obtain the following lower bound for the likelihood
\begin{align}
f_2\big( \{\vz_{(i)}\} | \vv, &\vc_i, N_i \big) \ge \overset{\sim}{f}_2\big( \{\vz_{(i)}\} | \vv, \vc_i, N_i \big) = \nn\\ & e^{\vv^T \left( \sum_{i=1}^P \mC_i \vzt_i \right) - \vv^T \left( \sum_{i=1}^P \frac{N_i}{2} \mC_i \mA \mC_i^T \right) \vv + const. } 
\label{eq:mult-like}
\end{align}
where $\vzt_i = \vz_i-N_i\left( \vp_{\vpsi_i} - \mA\vpsi_i \right)$, and $const.$ denotes the constant terms with respect to $\vv$. Assuming standard multivariate Gaussian $\cN(\vzero_{(D_2-1)K},\mI_{(D_2-1)K})$ prior for $\vv$ the lower bound for the complete-data likelihood is given by
\begin{align}
\label{eq:mult-post}
f_2\big( \{\vz_i\}, \vv | \vc_i, N_i \big) &\ge e^{-\frac{1}{2} (\vv-\vomega)^T \mOm^{-1} (\vv-\vomega) + const.},
\end{align}
where $\mOm = (\sum_{i=1}^P N_i \mC_i \mA \mC_i^T + \mI_{(D_2-1)K})^{-1}$ and $\vomega = \mOm \sum_{i=1}^P \mC_i \vzt_i$. From that lower bound we obtain an approximate posterior $\gt_{2,n}(\vv | \{\vz_i, \vc_i, N_i\}) = \cN(\vomega,\mOm)$. 
In the M-step, for computational efficiency, following the approach in \cite{Bohning88} we update the parameters by maximizing the expected lower bound 
$\Exp_{\gt_{2,n}}\big[ \log \overset{\sim}{f}_2\big( \{\vz_i\}, \vv | \vc_i, N_i \big) \big]$
instead of maximizing $\Exp_{\gt_{2,n}}\left[\log f_2\big( \{\vz_i\}, \vv | \vc_i, N_i \big) \right]$. It is shown in \cite{Bohning88} that maximizing the former is asypmtotically equivalent to, and computationally more efficient than maximizing the latter. 
Using this approximate posterior the M-step (i.e., parameter updates) of the variational EM algorithm can be derived as in the Gaussian case; however, the computational complexity might be prohibitive due to high dimensionality, e.g., $\mOm$ requires inverting a $(D_2-1)K \times (D_2-1)K$ matrix. Typically, the number of factors $K$ gets small values, whereas the number of categories may be large, e.g., a dictionary of words in topic modeling \cite{lda}. We next present a result that shows, indeed, the underlying dimensionality is $K \times K$.

\begin{pro}
\label{pro:mult}
At iteration $n$, the multinomial parameters $\vpsi_i$ and the coefficient vector $\vc_i$ can be updated 
as follows
\begin{align}
\vpsi_{i,n} &= \mPhi_n^T \vc_{i,n-1} \label{eq:mult-m}\\
\vc_{i,n} &= \arg\max_{\vc_i} -\frac{1}{2} \vc_i^T \mH_{i,n} \vc_i + \vc_i^T \vrho_{i,n}, \label{eq:c-m-mult}\\
\mH_{i,n} &= N_i \Big( \frac{(D_2-1)^2}{2D_2} \mF_n^{-1} + \frac{D_2-1}{2D_2} \mDel_n + \frac{1}{2} \mPhi_n\mPhi_n^T \nn \\  &~~~- \frac{1}{2D_2} (\mPhi_n\vone_{D_2-1})(\mPhi_n\vone_{D_2-1})^T \Big) \nn\\&~~~+ \sum_{j=1}^{D_1} \frac{\mB_{j,n}+\va_{j,n}\va_{j,n}^T}{\sigma_{ij,n}^2} \nn\\
\vrho_{i,n} &= \mPhi_n \vzt_{i,n} + \sum_{j=1}^{D_1} \frac{y_{ij}\va_{j,n}}{\sigma_{ij,n}^2}, \nn
\end{align}
using the matrices 
\begin{equation}
\begin{gathered}
\mF_n = \frac{1}{2} \mC_{n-1}\mN\mC_{n-1}^T + \mI_K \\
\mDel_n = \frac{\mI_K-\mF_n^{-1}}{D_2} \left[ \mF_n^{-1} + \left( \frac{\mF_n}{D_2-1}+\mI_K \right)^{-1} \hspace{-2mm} (\mI_K-\mF_n^{-1}) \right] \\
\mPhi_n = \mF_n^{-1} \mC_{n-1} \mZt_n + \mDel_n \mC_{n-1} (\mZt_n \vone_{D_2-1} \vone_{D_2-1}^T)
\end{gathered}
\label{eq:mult-e}
\end{equation}
where $\mB_{j,n}$ and $\va_{j,n}$ are given by \eqref{eq:gaus-e}, the $K \times (D_2-1)$ matrix $\mPhi = [\vphi_1 \cdots \vphi_{D_2-1}]$ is a reorganized form of the approximate posterior mean $\vomega = [\vphi_1^T \cdots \vphi_{D_2-1}^T]^T$, $\mN=\text{diag}(N_1,\ldots,N_P)$,
and $\mZt = [\vzt_1 \cdots \vzt_P]^T$, $\vzt_{i,n} = \vz_i-N_i\left( \vp_{\vpsi_{i,n}} - \mA\vpsi_{i,n} \right)$.
\end{pro}

\begin{IEEEproof}
See Appendix.
\end{IEEEproof}

Although not directly used in the learning algorithm, the complete form of the approximate posterior covariance (cf. \eqref{eq:mult-post}) is given by
$\mOm_n = \mI_{D_2-1} \otimes \mF_n^{-1} + \vone_{D_2-1}\vone_{D_2-1}^T \otimes \mDel_n$. 

\subsubsection{Coefficient Vector}

As shown in \eqref{eq:c-m-mult}, we have a quadratic programming problem for the coefficient vector $\vc_i$, which is simply solved as 
\be
\label{eq:c-m}
\vc_{i,n} = \mH_{i,n}^{-1} \vrho_{i,n}
\ee
unless there is a constraint on $\vc_i$, such as $c_{ik} \ge 0, \forall i,k$. If a constraint is added to the problem, a standard solver can be used, e.g., interior point methods. Additionally, depending on the application a convenient regularization, such as $L^1$-norm (Lasso) and $L^2$-norm (ridge regression), can be used to solve \eqref{eq:c-m}.

\subsubsection{Algorithm}

The resulting variational EM algorithm is summarized in Algorithm \ref{alg:gaus-mult}. Note that the EM steps for Gaussian (lines 4 and 5) and multinomial (lines 6 and 7) can be run in parallel, as shown in Fig. \ref{fig:em}. Moreover, the coefficient vectors $\{\vc_i\}$ can be updated in parallel (line 8).

\begin{algorithm}[h]
\caption{The proposed EM algorithm for the Gaussian-multinomial example}
\label{alg:em}
\baselineskip=0.5cm
  \begin{algorithmic}[1]
\STATE Input $\mY,\mZ$
    \STATE Initialize $\{\vc_i^{K \times 1},\sigma_{ij},\vpsi_i^{D_2-1 \times 1}\}, ~i=1,\ldots,P, ~j=1,\ldots,D_1,$    
    \WHILE{not converged}
    \STATE Compute Gaussian posterior parameters $\{\mB_j,\va_j\}$ as in \eqref{eq:gaus-e}
    \STATE Update Gaussian parameters $\{\sigma_{ij}^2\}$ as in \eqref{eq:gaus-m}
    \STATE Compute multinomial posterior parameters $\mF,\mDel,\mPhi$ as in \eqref{eq:mult-e}
    \STATE Update multinomial parameters $\{\vpsi_i\}$ as in \eqref{eq:mult-m}
    \STATE Update coefficients $\{\vc_i\}$ by solving \eqref{eq:c-m}
    \ENDWHILE
  \end{algorithmic}
\label{alg:gaus-mult}
\end{algorithm}

\subsection{Computational Complexity}
\label{sec:comp}

In the following theorem, we show that the computational complexity of Algorithm \ref{alg:gaus-mult} scales linearly with each dimension of the problem (i.e., number of instances, real-valued features, and categorical features). As a result, the proposed algorithm can be efficiently used for large datasets, as demonstrated in Section \ref{sec:exp}.
\begin{thm}
\label{thm:complex}
	At each iteration of the proposed EM algorithm, given by Algorithm \ref{alg:gaus-mult}, the computational complexity linearly scales with the number of instances $P$, the number of real-valued features $D_1$, and the number of categories $D_2$. Specifically, the complexity is given by $O(K^3P+K^2PD_1+KPD_2)$, where $K$ is the number of factors.	
\end{thm}

\begin{IEEEproof}
See Appendix.
\end{IEEEproof}

Note that the number of factors is not an input from data, but a design parameter typically chosen to be a small number compared to the number of instances, $K \ll P$. Furthermore, even in mildly complex datasets, it is also much smaller than the total number of features, $K \ll D_1+D_2$. Hence, in fact, $K$ can be dropped from the asymptotic complexity notation, which yields the following result. 
\begin{cor}
Algorithm \ref{alg:gaus-mult} scales with the data size as $O(P D)$, where $P$ is the number of instances and $D=D_1+D_2$ is the total number of features.
\end{cor}

\subsection{MSE Performance}

In this section, assuming the Gaussian and multinomial generative models are consistent with the observations (i.e., there is no model mismatch), we numerically compare the mean squared error (MSE), $\Exp[\|\vc_i-\vc_i\|^2]$, of Algorithm \ref{alg:gaus-mult} with the Cram\'{e}r-Rao lower bound (CRLB). With $\vu_j \sim \cN(\vmu_j,\mSig_j)$, the distribution of the Gaussian observations is $y_{ij} \sim \cN(\vc_i^T \vmu_j, \vc_i^T \mSig_j \vc_i + \sigma_{ij}^2)$.
It is straightforward to show that the Fisher information matrix of the Gaussian model is given by \cite[p. 47]{Kay93}
\begin{align}
\cF_g(\vc_i) = \sum_{j=1}^P \frac{\vmu_j \vmu_j^T}{\vc_i^T \mSig_j \vc_i + \sigma_{ij}^2} + 2 \frac{\mSig_j \vc_i \vc_i^T \mSig_j}{(\vc_i^T \mSig_j \vc_i + \sigma_{ij}^2)^2}, \nn
\end{align}
where the first and second terms are the contributions from the mean and variance, respectively.
On the other hand, in the multinomial model, due to the sum-exp term in the denominator of the likelihood, the Fisher information matrix is not tractable. Thus, we resort to numerical computation via Monte Carlo simulations as described next.

\begin{pro}
The Fisher information matrix $\cF_m(\vc_i)$ for the multinomial model is given by
\begin{align}
\label{eq:mult-fisher}
\cF_m& ( \vc_i) = \nn \\ &\Exp_{\vz_i}\left[ \frac{ \Exp_{\{\vv_j\}}\left[ p'(\vz_i | \{\vv_j\},\vc_i) \right] }{ \Exp_{\{\vv_j\}}\left[ p(\vz_i | \{\vv_j\},\vc_i) \right] } \left( \frac{ \Exp_{\{\vv_j\}}\left[ p'(\vz_i | \{\vv_j\},\vc_i) \right] }{ \Exp_{\{\vv_j\}}\left[ p(\vz_i | \{\vv_j\},\vc_i) \right] } \right)^T \right] 
\end{align}
\begin{align}
\text{where}~~ p'(\vz_i | \{\vv_j\},\vc_i) &= \frac{\partial}{\partial \vc_i} p(\vz_i | \{\vv_j\},\vc_i) \nn\\ &= p(\vz_i | \{\vv_j\},\vc_i) \sum_{j=1}^{D_2-1} \left( z_{ij}-N_i p_{ij} \right) \vv_j, \nn
\end{align}
and $p_{ij} = \frac{e^{\vc_i^T\vv_j}}{1+\sum_{j'=1}^{D_2-1} e^{\vc_i^T\vv_j'}}$ is the probability of category $j$ for instance $i$.
\end{pro}

\begin{IEEEproof}
The nontrivial part in the proof is justifying changing the order of expectation and differentiation. From the definition of Fisher information we have
\begin{align*}
\cF_m(\vc_i) &= \Exp_{\vz_i}\left[ \frac{\partial}{\partial \vc_i} \log p(\vz_i | \vc_i) \left( \frac{\partial}{\partial \vc_i} \log p(\vz_i | \vc_i) \right)^T \right] \\
&= \Exp_{\vz_i}\left[ \frac{ \frac{\partial}{\partial \vc_i} p(\vz_i | \vc_i) }{ p(\vz_i | \vc_i) } \left( \frac{ \frac{\partial}{\partial \vc_i} p(\vz_i | \vc_i) }{ p(\vz_i | \vc_i) } \right)^T \right] \\
&= \Exp_{\vz_i}\Bigg[ \frac{ \frac{\partial}{\partial \vc_i} \Exp_{\{\vv_j\}}\left[ p(\vz_i |  \{\vv_j\},\vc_i) \right] }{ \Exp_{\{\vv_j\}}\left[ p(\vz_i |  \{\vv_j\},\vc_i) \right] } \\ & ~~~~~~~~~~~~~~~~~ \Bigg( \frac{ \frac{\partial}{\partial \vc_i} \Exp_{\{\vv_j\}}\left[ p(\vz_i |  \{\vv_j\},\vc_i) \right] }{ \Exp_{\{\vv_j\}}\left[ p(\vz_i |  \{\vv_j\},\vc_i) \right] } \Bigg)^T \Bigg],
\end{align*}
where the differentiation can be brought into the expectation due to the Dominated Convergence Theorem \cite[p. 53]{Protter04} since expectation and differentiation are both limits, and $p(\vz_i |  \{\vv_j\},\vc_i)$ is a probability dominated by $1$. The derivative $p'(\vz_i | \{\vv_j\},\vc_i)$ directly follows from the likelihood 
$$p(\vz_i |  \{\vv_j\},\vc_i) = \frac{N_i!}{z_{i1}! \cdots z_{iD_2}!} \frac{ e^{\vc_i^T \sum_{j=1}^{D_2-1} z_{ij}\vv_j} }{\left( 1+\sum_{j'=1}^{D_2-1} e^{\vc_i^T\vv_j'} \right)^{N_i}}.$$
\end{IEEEproof}

The Fisher information expression given in \eqref{eq:mult-fisher} can be efficiently computed through Monte Carlo simulations, as shown in Algorithm \ref{alg:mult-fisher}. In Algorithm \ref{alg:mult-fisher}, a simplified notation is used by dropping some indices: $\vc$ is the coefficient vector to be estimated, $N$ is the number of multinomial experiments, $D$ is the number of categories, $R$ is the number of realizations to be averaged over, and $K$ is the number of factors. 

\begin{algorithm}[h]
\caption{Monte Carlo simulations for multinomial Fisher information}
\label{alg:em}
\baselineskip=0.5cm
  \begin{algorithmic}[1]
\STATE Input $\vc, N, D, R, K$
    \STATE Generate factor score matrices $\{\mV_r\}_{r=1,\ldots,R}$ with columns $\{\vv_{rd} \sim \cN(\vzero_K,\mI_K)\}_{d=1,\ldots,D-1}$
    \STATE Compute probability vectors \\
    $\Big\{\vp_r = \Big[ e^{\vc^T\vv_{r1}}/(1+\sum_{d=1}^{D-1}e^{\vc^T\vv_{rd}}) \cdots$ \\ $~~~~~~~~~~~~~~~~~~~~~~~~~~~~~~~~~~1/(1+\sum_{d=1}^{D-1}e^{\vc^T\vv_{rd}}) \Big] \Big\}_i$
    \FOR{r=1,\ldots,R}
    \STATE Generate observation vector $\vz_r \sim \cM(N,\vp_r)$
    \STATE Compute likelihoods $\vell_r = [\ell_{r1} \cdots \ell_{rR}]^T$ with $\ell_{rs} = \cM(\vz_r | N,\vp_s)$
    \STATE Compute average likelihood $\bar{\ell}_r = \left( \ell_{r1} +\cdots+ \ell_{rR} \right)/R$
    \STATE Compute matrix $\mLam_r = [\vlam_{r1} \cdots \vlam_{rR}]$ where $\vlam_{rs} = \mV_s \left( \vz_r-N\vp_s \right)$
    \STATE Compute average derivative $\bar{\vell'}_r = \mLam_r \vell_r / R$
    \ENDFOR
    \STATE Compute $\cF_m = \frac{1}{R} \sum_{r=1}^R \frac{\bar{\vell'}_r \bar{\vell'}_r^T}{\bar{\ell}_r^2}$
  \end{algorithmic}
\label{alg:mult-fisher}
\end{algorithm}

The overall Fisher information of the multimodal model and CRLB are given by 
\begin{align*}
\cF(\vc_i) &= \cF_g(\vc_i)+\cF_m(\vc_i) \\
\Exp[\|\vc_i-\vc_i\|^2] &\ge \text{trace}(\cF(\vc_i)^{-1}).
\end{align*}

\begin{figure}
\centering \includegraphics[width=\linewidth]{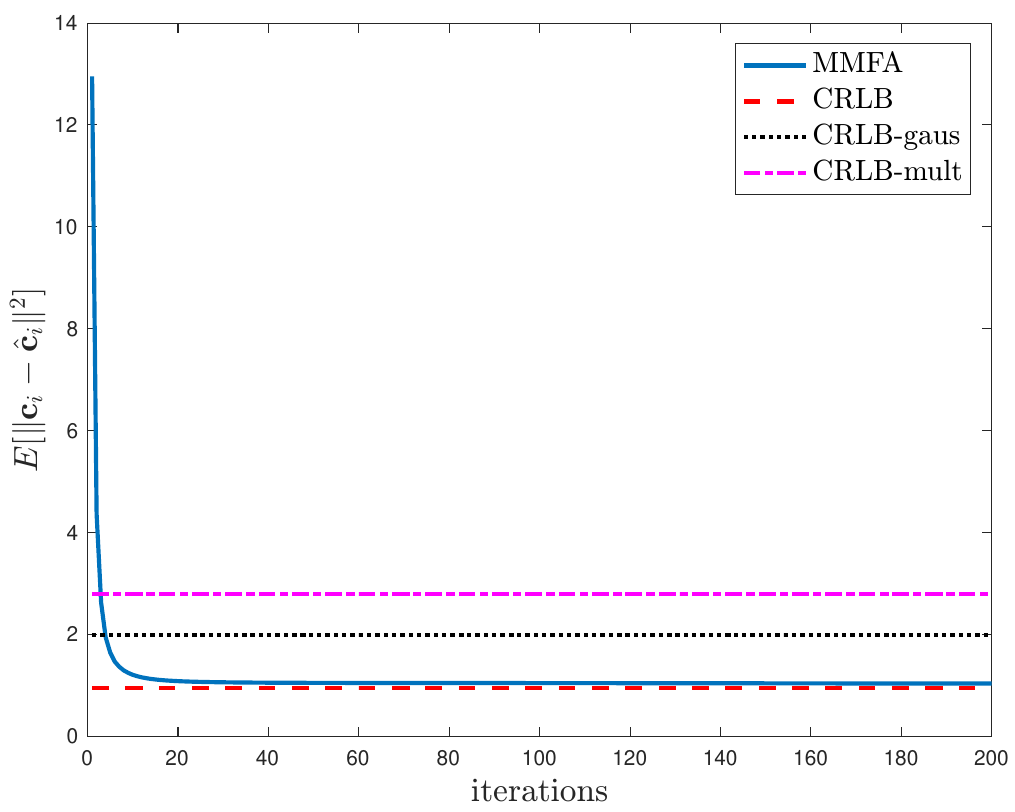}
\caption{MSE performance of the proposed variational EM algorithm. It converges to the multimodal bound CRLB quickly in 20 iterations. The Gaussian and multinomial components of CRLB are also shown.}
\label{fig:mse}
\end{figure}

We next present simulation results for the MSE performance. In the simulated data, the number of Gaussian features and the number of multinomial categories are $D_1=D_2=5$, the number of multinomial experiments is $N_i=40$, and the number of instances is $P=100$. In the MMFA algorithm, the number of factors is $K=3$, ridge regression is used for updating $\vc_i$ (see \eqref{eq:c-m}) with weight $10^{-6}$, and the hyperparameters $\alpha=1$ and $\beta=0.1$ are used for the inverse-gamma prior of $\sigma_{ij}^2$ (see \eqref{eq:gaus-m}). The statistical expectation in the multinomial Fisher information is computed by averaging over $R=2000$ realizations (see Algorithm \ref{alg:mult-fisher}).
Fig. \ref{fig:mse} shows that MSE of the proposed MMFA algorithm converges close to the CRLB (red dashed line) as early as in $20$ iterations. The CRLB for Gaussian data, $\text{trace}(\cF_g(\vc_i)^{-1})$, and multinomial data, $\text{trace}(\cF_m(\vc_i)^{-1})$, are also shown in the same figure with black dotted line and purple dashed line, respectively.

\begin{figure}
\centering \includegraphics[width=0.9\linewidth]{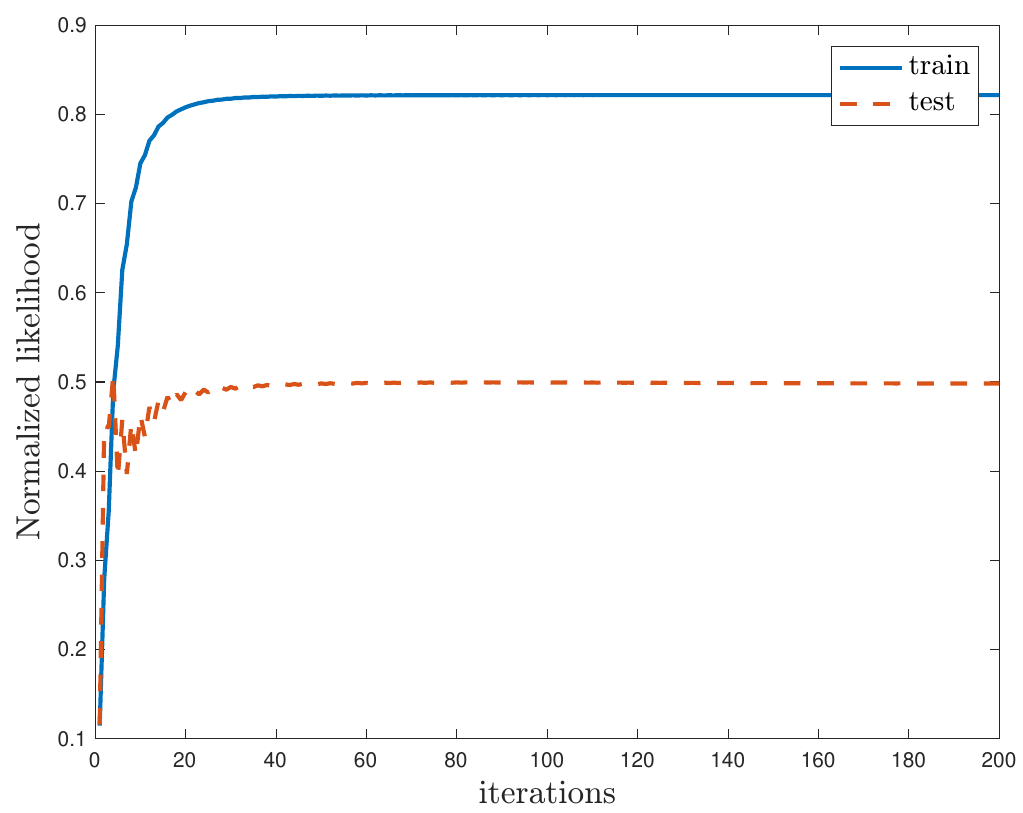}
\caption{Likelihood under the proposed model for training and test data normalized by the likelihood under the true model.}
\label{fig:like}
\end{figure}

Using the same simulation setup we show in Fig. \ref{fig:like} that under the MMFA model the likelihood of both the training ($P=100$) and the unseen test data ($P=10$) increase with the iterations and appear to converge to a limit. On the vertical axis, the likelihood of the multimodal data under the estimated model normalized by the likelihood under the true model is shown, i.e., 
$\frac{p(\{\vy_i\} | \{\vc_i,\va_j, \hat{\sigma}_{ij}^2 \}) p(\{\vz_i\} |  \{\vc_i\}, \mPhi)}{p(\{\vy_i\} | \{\vc_i,\vu_j, \sigma_{ij}^2 \}) p(\{\vz_i\} |  \{\vc_i, \vv_j\})}.$

\section{Experiments}
\label{sec:exp}

In this section, we will demonstrate the power of MMFA in large datasets for different tasks, such as generalization to unseen data, anomaly detection, 
data imputation, and recommender systems. We start with the New York City (NYC) Taxi dataset \cite{NYC}, and conclude with the MovieLens dataset. The codes used to produce the results in this paper are publicly available \footnote{\url{https://github.com/maktukmak/MMFA}}.

\subsection{NYC Taxi Data}

This dataset provides trip records for the yellow and green taxicabs, and the for-hire vehicles in NYC. Here we use the data from yellow taxis from February 2019 \cite{NYC}. The dataset includes, for each recorded trip, the pick-up and drop-off dates, times and locations, trip distances, itemized fares, rate types, payment types, tip amounts, and passenger counts. 
The considered dataset has almost $7$ million trip records. We first extracted a subset of the variables in the dataset, and filtered them to reduce bias. Specifically, we only considered trips with credit card payments since in most of the trips with cash payment the tip amount is unrealistically recorded as zero. We also disregarded trips that report fewer than $1$ or more than $6$ passengers (which is the legal limit). Location is reported in terms of the taxi zone id from $1$ to $263$. Unknown locations, denoted by the id $264$ or $265$, are ignored. Finally, we removed trips reporting a trip distance smaller than $0.1$ mile or greater than $40$ miles, and fare amounts less than $\$1$ and more than $\$200$. After preprocessing, the data size decreased to around $P=6.4 \times 10^6$. The considered features consist of numerical ones, namely the tip amount, fare amount, number of passengers, and trip distance, which are modeled using a four-dimensional Gaussian distribution ($D_1=4$), and categorical ones, namely the pick-up day ($D_{2,1}=7$ choices), pick-up time ($D_{2,2}=24$ choices), and location ($D_{2,3}=263$ choices). Categorical distribution with one-of-K (a.k.a. one-hot) representation yields a total number of $298$ features. For all categorical features, the number of trials is $N_i=1, \forall i$. 

\begin{figure}
\centering \includegraphics[width=1\linewidth]{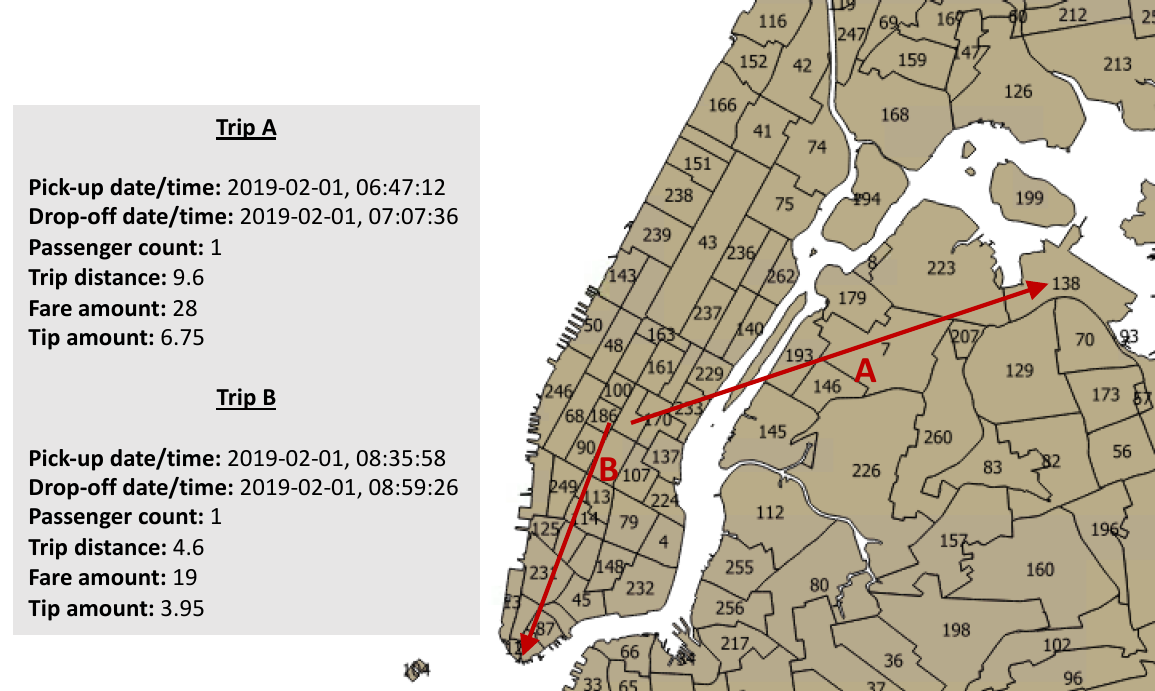}
\caption{Sample trips from the NYC Taxi dataset. Trip A is from Midtown Manhattan to LaGuardia Airport, and trip B is from Penn Station to Financial District.}
\label{fig:trip}
\end{figure}

Note that the numerical and categorical features arise from the same physical event (i.e., taxi trip), and hence they are dependent in general. For instance, it is seen that the tip amount, fare amount, and the number of passengers statistically depend on the pick-up and drop-off locations, e.g., trips from and to Manhattan statistically have higher tip percentages. Two sample trips are shown on the NYC map in Fig. \ref{fig:trip}.

\begin{figure}
\centering 
\includegraphics[width=.9\linewidth]{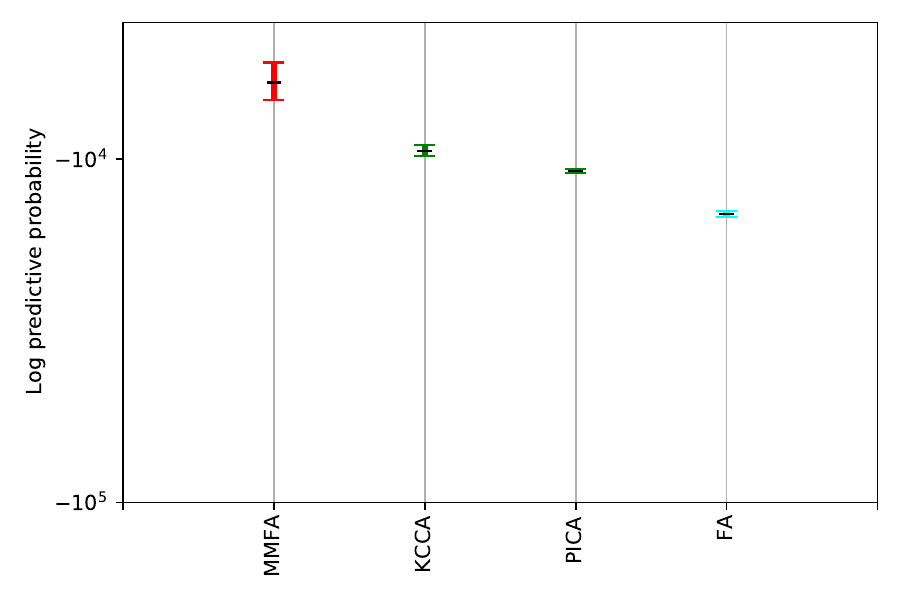}
\caption{Generalization performance comparison between the proposed MMFA algorithm and existing data fusion methods which treat categorical data as numerical. The mean and $95\%$ confidence intervals are shown for log predictive likelihood.
Higher values indicate higher generalization performance.}
\label{fig:perp}
\end{figure}

\textbf{\textit{Generalization performance:}}  For generative models  a common performance metric, the predictive log-likelihood value,  is often used to assess the ability of the model to generalize to unseen data in training \cite{lda}. The predictive likelihood value is computed for an unseen data instance as follows. The trained MMFA model produces posterior distributions from estimated model parameters obtained using successive E and M steps $\{$(M-step)$_m \}_{m=1}^M$ shown in Fig. \ref{fig:em}.  The final M-step, (M-step)$_c$ in Fig. \ref{fig:em}, uses these parameters to compute the factor score vector $\vc_i$ associated with an unseen data instance $\vx_i$, which specifies the likelihood function through the generative model shown in Fig. \ref{fig:oper}.

We compare the proposed MMFA algorithm with Kernel Canonical Correlation Analysis (KCCA) \cite{Bilenko}, Parallel Independent Component Analysis (PICA) \cite{Liu09,Pearlson15}, and the standard factor analysis (FA) method \cite{Murphy12} on the NYC Taxi dataset.
As compared to standard CCA, KCCA algorithm enables fusion of data with multiple modalities by choosing proper kernels. Here, we use a linear kernel for the numerical modalities to form the symmetric Gram matrix. For the categorical data, we compute the Gram matrix by using the Hamming distances between observations, which is a proper similarity measure for the binary-coded categorical variables. Canonical components are then computed by projecting the Gram matrices to a lower dimensional latent space through demixing matrices for each modality. The KCCA objective maximizes the correlation of each of these canonical components across the modalities. This is usually performed by solving a generalized eigenvalue problem.
On the other hand, PICA assumes non-Gaussian independent components for each modality. Here, we use a non-quadratic exponential decay function for the log of the probability distribution functions of the components as suggested in \cite{Hyvarinen00}. The modalities are correlated through the mixing matrices instead of the latent components. To implement the cross-modality optimization in PICA, a term is added to the PICA objective that encourages maximization of the correlation between the components of the matrices. Centering and whitening are applied in advance as a preprocessing step. 
Conversely, FA assumes spherical Gaussian components for the latent variables, i.e., factor loading coefficients. It is a generalization of PCA, where the noise covariance matrix is diagonal and has $D$ free parameters. Also, the orthonormality constraint on the factor loading matrix is relaxed in FA models. The model is usually fit by using the EM algorithm, where the E-step computes posterior distributions of the latent variables, and the M-step computes point estimate for the factor loading matrix. We adopted the FA model as a representative of the class of fusion algorithms that treat all data modalities as numerical by  concatenating them to form a long feature vector in a linear model. 

MMFA achieves fusion by generating common latent factor scores to generate both modalities under a Bayesian graphical model. Under this model, the data modalities are conditionally independent given the factor loading coefficients, allowing 
modality-specific probabilistic models to be fused together. As generalized linear models for a large variety of data types and distributions are available (Table \ref{tab:exp}), the MMFA can be applied to a wide range of data types beyond the Gaussian/multinomial case considered here.  
In contrast, Kernel CCA fuses data types by performing an eigendecomposition on a distance matrix or, equivalently, a similarity matrix.  Instead of incorporating explicit probability distribution models for different data types, as in MMFA, KCCA accounts for different data types through transformation of the similarity matrix via kernelization, where the type of kernel is selected to match the data type.    
As KCCA performs an eigendecomposition of a data similarity matrix its computational complexity is of order $O(P^3)$ as compared to only order $O(PD)$ for FA and MMFA.
On the other hand, PICA jointly models the modalities by maximizing the correlation between the components of the mixing matrices. If we denote dimension of the two modalities $D_1$ and $D_2$, there are $D_1 \times D_2$ possible pairs to compute the correlation. Indeed, the algorithm chooses single pair at each iteration, which has the maximum correlation as compared to the other pairs. Hence, it is not straightforward how to extend this model to more than two modalities in an efficient way. Also, note that the categorical data is still modeled as numerical data.

We compute the log predictive marginal likelihood on the test set to compare the models. To this end, the latent variables are integrated out. Particularly, for MMFA, the Gaussian parameters $\mathbf{u}_j$ and multinomial parameters $\mathbf{v}_j$ are integrated out. For FA, PICA and KCCA, the latent variables are assigned per data point as opposed to MMFA, which are integrated out in a closed form. Note that KCCA likelihoods are computed on the higher dimensional kernel space instead of the observation space.

{A train-validate-test split with $(0.6, 0.2, 0.2)$ ratio is applied to the data. 
Using BIC, the best number of latent dimensions is found to be $10$ for MMFA ($K=10$), $6$ for FA, $8$ for KCCA. The BIC value $k \log(P)-2\log(p(\mX))$ penalizes the negative log-likelihood score with the number of model parameters $k$, hence the $K$ value with the smallest BIC value is selected for each algorithm.
Figure \ref{fig:perp} shows, for MMFA, FA, and KCCA, the mean and $95\%$ confidence interval of the log predictive marginal likelihood values. The random train-validate-test split was repeated 20 times to compute the mean and confidence interval of the log predictive likelihoods. The mean values for MMFA, KCCA, PICA, and FA are -5982, -9471, -10833, -14429, respectively. 
By modeling the categorical data appropriately and fusing it with numerical data with probabilistic models, MMFA achieves much better generalization performance compared to other data fusion techniques that treat categorical data the same way as they do with numerical data.}

\ignore{
\begin{table}[h!]
\caption{Perplexity values}
\centering
\begin{tabular}{|c|c|c|c|c|}
\hline
& jICA & IVA & FA & MMFA \\
\hline
Perplexity & 1511 & 444 & 411 & 77 \\
\hline 
\end{tabular}
\label{tab:perp}
\end{table}}

\begin{table}[t]
\caption{Some anomalies found by MMFA in the test set.}
\centering
\begin{tabular}{|c|c|c|c|c|c|}
\hline
Location& Day & Hour & Tip & Fare & Passenger \\
\hline
132 & 6 & 4 & 90.00 & 46.00 & 1 \\
66& 2 & 10 & 29.30 & 26.00 & 1 \\
91 & 2 & 7 & 0.00 & 59.00 & 1 \\
237 & 2 & 23 & 2.70 & 59.50 & 1 \\
79 & 5 & 1 & 5.00 & 63.50 & 2 \\
\hline 
\end{tabular}
\label{tab:anom}
\end{table}

\textbf{\textit{Anomaly detection:}} We next demonstrate the anomaly detection performance of MMFA on the NYC taxi data. We first fit the MMFA model on the training set, and then sort the likelihoods of instances in the validation set with respect to the trained model. Finally, in the test, we compare the likelihood of each instance with the likelihoods of validation set, and declare anomalous if it is smaller than the $(1-\delta)\%$ of the validation likelihoods, where $\delta$ is a small number representing the statistical significance level. For $\delta=0.05$, the top five anomalies with the smallest likelihoods are shown in Table \ref{tab:anom}. The first three anomalies are obvious as the tip/fare ratio is unexpectedly high or low. However, the last two anomalies in the table can be considered as interesting findings of the MMFA model. The reason why they appear among the top anomalies is not only the tip/fare ratio, but in fact the inconsistency between the location and the tip percentage, $4.5\%$ and $7.9\%$, respectively. Their locations are both in Manhattan with tip percentage mean and standard deviation of $(27.3\%,10.1\%)$ for location 237 and $(25.2\%,10.4\%)$ for location 79. While these trips are detected by MMFA as anomalous with tip percentages $4.5\%$ and $7.9\%$, there are other trips with smaller tip percentages that are deemed nominal in locations with smaller mean tip percentages, e.g., $5.1\%$ in location 132 (Queens) where mean and standard deviation is $(19\%, 10.7\%)$. Since there is no ground truth (i.e., nominal and anomalous labels) in the dataset, such comparative evaluation is useful in showing MMFA's success in anomaly detection. Note also that MMFA is a completely unsupervised algorithm.

\subsection{MovieLens Data}

Our next application is recommender systems, in which the objective is to learn user patterns and provide successful item recommendations to the users. While the commonly used collaborative filtering techniques in recommender systems typically use only the interactions between the users and items to learn the user patterns, there are also hybrid methods that combine user-item interactions with side information, such as user demographics and item features, for better recommendation performance \cite{Aggarwal16}. However, the existing hybrid methods mainly convert the categorical side information, such as gender, occupation, item genre, country, etc., to numerical data for fusing multimodal data. 

The MovieLens dataset, which is commonly used as a benchmark dataset in recommender system applications, has three different versions, 100K, 1M, and 10M, in terms of the number of interactions between the users and the movies, i.e., user ratings for items. 
To show the scalability of the proposed MMFA algorithm, we use the MovieLens-10M dataset, which has a little more than 10M ratings from 71567 users to 10681 items. The size and sparsity of this dataset, where $98.7\%$ of $71567\times10681$ possible interactions is missing, brings significant challenges. In addition to the ratings, two item side information, release date and genre, are available in the dataset. We model release date using a univariate Gaussian distribution, and each of 21 genre categories using a Bernoulli (i.e., binary categorical) distribution since an item can have multiple genres.

We train the MMFA model on heterogeneous data from $P=10681$ items, consisting of numerical ratings from $D_{1,1}=71567$ users, release date ($D_{1,2}=1$), and categorical genre information ($D_{2,j}=2$, $j=1,\ldots,21$ with $N_{i,j}=1$ trial $\forall i,j$), to find the latent vector $\vc_i$ for each item $i$. 
The number of factors $K$ is chosen as 10 using BIC. 
In latent variable models for collaborative filtering, such as probabilistic matrix factorization (PMF), the rating $r_{ij}$ of item $i$ from user $j$ is commonly modeled using a Gaussian distribution $\cN(\vc_i^T \vu_j, \sigma^2)$, where $\vu_j$ is a latent factor score vector representing user $j$, and $\sigma^2$ is the variance parameter. 

Following the recommender systems literature we consider two experiment setups called warm start and cold start. In warm start, each user or each item has at least one rating in the training set, whereas in cold start, for some users or items the recommender system has to completely rely on side information as there was no related rating in training. For warm start, $60\%-20\%-20\%$ training-validation-test split is used for the ratings of each item. On the other hand, for cold start, all the ratings of $20\%$ of the items are used in the test set, and similarly $20\%$ in the validation set. 

\textbf{\textit{Data imputation:}} We first evaluate MMFA for predicting ratings in terms of MSE, $\frac{1}{|\sO|} \sum_{i,j \in \sO} (r_{ij}-\vc_{i,n}^T \va_{j,n})^2$, where $\va_{j,n}$ is the posterior mean of $\vu_j$ (see \eqref{eq:gaus-e}), $\sO$ is the set of observed ratings in the test and $|\sO|$ is its cardinality. MMFA is compared with several collaborative filtering algorithms, namely SVDpp, NMF, PMF, and BPMF. Specifically, the SVDpp method \cite{Koren08} performs a matrix factorization on the rating matrix including implicit ratings to find the user and item matrices. The nonnegative matrix factorization (NMF) algorithm, similar to the singular value decomposition (SVD), computes a matrix factorization, but by enforcing both user and item matrices to be nonnegative \cite{Luo14}. PMF also applies a matrix factorization on the rating matrix by assuming Gaussian latent variables for both users and items \cite{Mnih08}. In Bayesian PMF (BPMF), additional prior distributions are assumed for the hyperparameters of user and item latent  variables \cite{Salakhutdinov08}. All of the four benchmark models use only the rating matrix without any side information. Leveraging item side information MMFA has a clear advantage over them in the cold-start setting, hence we only compare the algorithms in the warm-start setting by averaging over 10 experiments with random training-validation-test split. As shown in Table \ref{tab:imput}, MMFA achieves the best MSE performance also in the warm-start setting by utilizing the item side information available in the dataset. 

\begin{table}[t]
\caption{MSE on MovieLens-10M dataset with warm-start.}
\centering
\begin{tabular}{|c|c|c|c|c|c|}
\hline
& SVDpp & NMF & PMF & BPMF & {\bf MMFA} \\
\hline
MSE & 0.659 & 0.766 & 0.691 & 0.671 & {\bf 0.632} \\
\hline 
\end{tabular}
\label{tab:imput}
\end{table}

\textbf{\textit{Recommendation accuracy:}} Finally, we evaluate MMFA's performance in terms of the accuracy of recommended movies to the users. For movie datasets, recall (i.e., true positive rate) is computed as the ratio ``number of movies user liked in the recommendations $/$ total number of movies user liked". In Table \ref{tab:recall}, for 10 recommendations, we report the recall averaged over all test users and 10 different experiments with random splits in both warm- and cold-start settings. Here we compare the proposed MMFA approach with state-of-the-art recommender systems that are capable of utilizing side information. Among the considered state-of-the-art methods, LCE \cite{Saveski14}, DecRec \cite{Barjasteh16}, and KMF \cite{Zhou12kernel} could not scale well to the MovieLens-10M dataset due to the memory limitations. These algorithms store similarity matrices for users and items based on the side information, causing $O(P^2+D^2)$ space complexity, where $P$ and $D$ are the number users and items. Moreover, the KMF algorithm inverts such matrices, increasing its space complexity to $O(P^3+D^3)$. In the MovieLens-10M dataset, only items have side information, and $D=10681$. On the other hand, LightFM \cite{Kula15}, which incorporates the side information into the rating matrix and performs matrix factorization on the enhanced data in a non-probabilistic way, scales well to the MovieLens-10M data. The superior performance of MMFA can be attributed to its natural handling of data fusion through appropriate probabilistic models while LightFM embeds categorical features into numerical values for data fusion. 

\begin{table}[h]
\caption{Recall with 10 recommendations on MovieLens-10M dataset with warm- and cold-start.}
\centering
\begin{tabular}{|c|c|c|}
\hline
Recall & LightFM & {\bf MMFA} \\
\hline
Warm-start & 0.886 & {\bf 0.902} \\
Cold-start & 0.824 & {\bf 0.886} \\
\hline 
\end{tabular}
\label{tab:recall}
\end{table}

\section{Conclusion}
\label{sec:conc}

A general unsupervised Bayesian framework based on the exponential family was proposed for the joint analysis of heterogeneous datasets. The proposed model, called Multimodal Factor Analysis (MMFA), uses the most appropriate probability distribution from the exponential family for each data modality, and fuses them by modeling their natural parameters through a common latent vector for each instance. To fit the model on large high-dimensional datasets, we proposed a computationally efficient variational Expectation-Maximization (EM) algorithm, which scales linearly with the number of features and the number of instances. On the common real-valued and categorical data combination, we showed that the algorithm quickly converges to the Cramer-Rao Lower Bound (CRLB) when there is no model mismatch. The proposed algorithm was also evaluated on two high-dimensional and heterogeneous datasets, the NYC Taxi dataset and the MovieLens-10M dataset, for various machine learning tasks. Specifically, the experiments demonstrated that the proposed MMFA model generalizes to unseen data better than the state-of-the-art data fusion techniques such as KCCA and PICA, provides meaningful anomaly detection results, predicts missing data better than the collaborative filtering techniques, and gives more accurate recommendations than the state-of-the-art recommender systems. We should note here that despite our efforts for a fair comparison between algorithms, the benchmark algorithms could possibly be further optimized to improve their performances. As future work, we plan to investigate (i) deep versions of MMFA which fuses different modalities in lower levels of hyper-parameters in a hierarchical Bayesian setup, (ii) links and comparisons with generative neural network models, such as Variational Autoencoders, Restricted Boltzmann Machines (RBM), Generative Adversarial Networks (GAN), and (iii) stochastic optimization methods for variational EM to improve further the memory complexity for extremely large datasets.

\section*{Appendix}

\subsection*{Proof of Proposition \ref{pro:mult}:}

For notational simplicity, we will drop the iteration index $n$ in the E-step. Defining the diagonal matrix $\mN=\text{diag}(N_1,\ldots,N_P)$ we start with manipulating $\mOm$ using \eqref{eq:lse} and \eqref{eq:mult-post},
\begin{align*}
\mOm &= \left(\sum_{i=1}^P N_i \mC_i \mA \mC_i^T + \mI_{(D_2-1)K}\right)^{-1} \\
&= \left(\mA \otimes \mC\mN\mC^T + \mI_{(D_2-1)K}\right)^{-1} \\
&= \Big[ \mI_{D_2-1} \otimes \Big( \frac{1}{2}\mC\mN\mC^T + \mI_{K} \Big) \\ &~~~~+ \Big(\vone_{D_2-1} \otimes \mC \Big) \Big(-\frac{\mN}{2D_2} \Big) \Big(\vone_{D_2-1}^T \otimes \mC^T \Big) \Big]^{-1}.
\end{align*}
Using the Matrix Inversion Lemma we can write
\begin{align*}
\mOm &= \mI_{D_2-1} \otimes \mF^{-1} - \vone_{D_2-1} \otimes \mF^{-1} \mC \\ & \underbrace{\left[ -2D_2 \mN^{-1} + (D_2-1) \mC^T \mF^{-1} \mC \right]^{-1}}_{\mGam} \vone_{D_2-1}^T \otimes \mC^T\mF^{-1}  \\
&= \mI_{D_2-1} \otimes \mF^{-1} - \vone_{D_2-1} \vone_{D_2-1}^T \otimes \mF^{-1} \mC \mGam \mC^T\mF^{-1},
\end{align*}
where $\mF = \frac{1}{2} \mC \mN \mC^T + \mI_K$. Using again the Matrix Inversion Lemma for $\mGam$ we obtain
\begin{align*}
\mOm &= \mI_{D_2-1} \otimes \mF^{-1} + \vone_{D_2-1}\vone_{D_2-1}^T \otimes \mDel,
\end{align*}
where 

$\mDel = \mF^{-1} \mG \left[ \mI_K + (D_2-1) \left( \mG+\mI_K \right)^{-1} \mG \right] \mF^{-1}$ and $\mG = \frac{1}{2D_2} \mC\mN\mC^T = \frac{\mF-\mI_K}{D_2}$.

We continue with putting the mean $\vomega$ of the approximate posterior in a compact and computationally efficient form. 
\begin{align*}
\vomega &= \mOm \sum_{i=1}^P \mC_i \vzt_i = \mOm \left[ \sum_{i=1}^P \zt_{i1} \vc_i^T \cdots \sum_{i=1}^P \zt_{iD_2-1} \vc_i^T \right]^T \\ &= \mOm \left[ (\mC\vxi_1)^T \cdots (\mC\vxi_{D_2-1})^T \right]^T \\
&= \Bigg[ \left(\mF^{-1} \mC\vxi_1 + \mDel \mC \sum_{j=1}^{D_2-1} \vxi_j \right)^T \cdots \\ &~~~~~~~~~~~~~~~~~~~~~~\left(\mF^{-1} \mC\vxi_{D_2-1} + \mDel \mC \sum_{j=1}^{D_2-1} \vxi_j \right)^T \Bigg]^T,
\end{align*}
where $\vxi_j = [\zt_{1j} \cdots \zt_{Pj}]^T, j=1,\ldots,D_2-1$, and we used the $\mOm$ expression derived above. Reorganizing the vector $\vomega = [\vphi_1^T \cdots \vphi_{D_2-1}^T]^T$ as the matrix $\mPhi = [\vphi_1 \cdots \vphi_{D_2-1}]$, in a more compact form, we can write
\begin{align*}
\mPhi = \mF^{-1} \mC \mZt + \mDel \mC (\mZt \vone_{D_2-1} \vone_{D_2-1}^T),
\end{align*}
where $\mZt = [\vxi_1 \cdots \vxi_{D_2-1}] = [\vzt_1 \cdots \vzt_P]^T$.

In the M-step, the update equation for $\vpsi_i$ at iteration $n$ is found from \eqref{eq:lse} as,
\begin{align*}
\vpsi_{i,n} &= \arg\max_{\vpsi_i} \Exp_{\cN\left(\vv|\vomega,\mOm\right)}\Big[ - \frac{1}{2} (\veta_i-\vpsi_i)^T \mA (\veta_i-\vpsi_i) \\ &~~~~- (\veta_i-\vpsi_i)^T \nabla \lse(\vpsi_i) - \lse(\vpsi_i) \Big] \\
0 &= \Exp_{\cN\left(\vv|\vomega,\mOm\right)}\left[ \mA (\mV^T\vc_i-\vpsi_{i,n}) + \nabla \lse(\vpsi_i) - \nabla \lse(\vpsi_i) \right]\\
\vpsi_{i,n} &= \mPhi_n^T\vc_{i,n-1}
\end{align*}
For the update of the coefficient vector estimate $\vc_{i,n}$ (cf. \eqref{eq:c-m-gaus}), from \eqref{eq:mult-like}, the part related to the multinomial model is given by
\begin{align*}
\zeta_n(\vc_i) &= -\frac{N_i}{2} \Exp_{\cN\left(\vv|\vomega,\mOm\right)}\left[ \vv^T \mC_i \mA \mC_i^T \vv \right] + \vomega_n^T \mC_i \vzt_i \\
&= -\frac{N_i}{2} \Tr\left( \mC_i \mA \mC_i^T (\mOm_n+\vomega_n\vomega_n^T) \right) + \vc_i^T \mPhi_n \vzt_i \\
&= -\frac{N_i}{2} \Tr\left( (\mI_{D_2-1} \otimes \vc_i) \mA (\mI_{D_2-1} \otimes \vc_i^T) (\mOm_n+\vomega_n\vomega_n^T) \right) \\&~~~+ \vc_i^T \mPhi_n \vzt_i \\
&= -\frac{N_i}{2} \Tr\left( (\mA \otimes \vc_i\vc_i^T) (\mOm_n+\vomega_n\vomega_n^T) \right) + \vc_i^T \mPhi_n \vzt_i \\
&= -\frac{N_i}{2} \Tr\left( \mA \otimes \vc_i\vc_i^T \mF_n^{-1} + \mA\vone_{D_2-1}\vone_{D_2-1}^T \otimes \vc_i\vc_i^T \mDel_n \right) \\&~~~-\frac{N_i}{2} \vomega_n^T (\mA \otimes \vc_i\vc_i^T) \vomega_n + \vc_i^T \mPhi_n \vzt_i \\
&= -\frac{N_i}{2} \vc_i^T \left( \Tr(\mA) \mF_n^{-1} + \Tr(\mA\vone\vone^T)\mDel_n \right) \vc_i \\&~~~-\frac{N_i}{2} \sum_{j=1}^{D_2-1} \vomega_{j,n}^T \vc_i\vc_i^T \vomega_{j,n} + \vc_i^T \mPhi_n \vzt_i  \\&~~~-\frac{N_i}{2D_2} \sum_{j=1}^{D_2-1} \sum_{j'=1}^{D_2-1} \vomega_{j,n}^T \vc_i\vc_i^T \vomega_{j',n} \\
&= -\frac{N_i}{2} \vc_i^T \Big( \frac{(D_2-1)^2}{2D_2} \mF_n^{-1} + \frac{D_2-1}{2D_2} \mDel_n + \frac{1}{2} \mPhi_n\mPhi_n^T \\&~~~- \frac{1}{2D_2} (\mPhi_n\vone)(\mPhi_n\vone)^T \Big) \vc_i^T + \vc_i^T \mPhi_n \vzt_i,
\end{align*}
which concludes the proof.

\subsection*{Proof of Theorem \ref{thm:complex}:}

\ignore{
We start with the vMF E-step (line 4 in Algorithm \ref{alg:em}). The most expensive computation in the vMF E-step is the posterior mean direction $\vb_k$, given by \eqref{eq:vmf_E_mean}. Note that the sum of geolocation vectors $\sum_{n=1}^{N_i} \vw_{in}$ is computed offline once for each hashtag $i$; hence the number of geotagged tweets $N_i$ does not contribute to the computational complexity. Each $\vb_k$ has a computational complexity of $O(P)$. As a result, the computational complexity for the vMF E-Step is $O(KP)$. 
There is no expensive computation in the vMF M-step (line 5 in Algorithm \ref{alg:em}). \newline \indent
}

In the Gaussian E-step (line 4 in Algorithm \ref{alg:em}), the most expensive computations are $\mC \mSig_j^{-1} \mC^T$ and $\mB_j \mC$ for each feature $j$, resulting in $O(K^2 P D_1)$ computations. The matrix inversion for all $j$ is $O(K^3 D_1)$, but this is cheaper than $O(K^2 P D_1)$ since $K \ll P$. Similarly in the Gaussian M-step (line 5), computing $\vc_i^T \mB_j$ for each $i,j$ pair yields $O(K^2 P D_1)$ complexity.
In the multinomial model, both $\mC \mZt$ in the E-step (line 6) and $\mPhi^T \vc_i$, for $i=1,\ldots,P$, in the M-step (line 7) have a complexity of $O(KPD_2)$, and the rest of the computations are cheaper. 
Finally, for the coefficients $\vc_i$ (line 8) we solve a quadratic program for each $i$. In solving a possibly constrained quadratic program for each $\vc_i$, the number of iterations, in practice, is bounded by a constant; and in each iteration, linear algebra operations in the $K$-dimensional space are performed. Hence, the overall complexity for solving the quadratic programs is $O(K^3P)$. Note also that each $\vc_i$ can be updated in parallel. The computation of $\{\mH_i\}$ and $\{\vrho_i\}$ have $O(K^2(K+P+D_2+P D_1))$ and $O(KP(D_1+D_2))$ complexity, respectively. Combining all the complexities we get $O(K^3P+K^2PD_1+KPD_2)$.

\begin{IEEEbiography}
[{\includegraphics[width=1in,height=1.4in,clip,keepaspectratio]{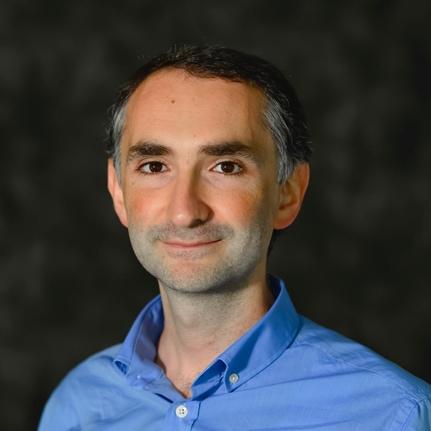}}]{Yasin Yilmaz}
(S'11-M'14-SM'20) received the Ph.D. degree in Electrical Engineering from Columbia University, New York, NY, in 2014. He is currently an Assistant Professor of Electrical Engineering at the University of South Florida, Tampa. His research intere ts include statistical signal processing, machine learning, and their applications to computer vision, cybersecurity, IoT networks, energy systems, transportation systems, and communication systems. 
\end{IEEEbiography}


\begin{IEEEbiography}
[{\includegraphics[width=1in,height=1.4in,clip,keepaspectratio]{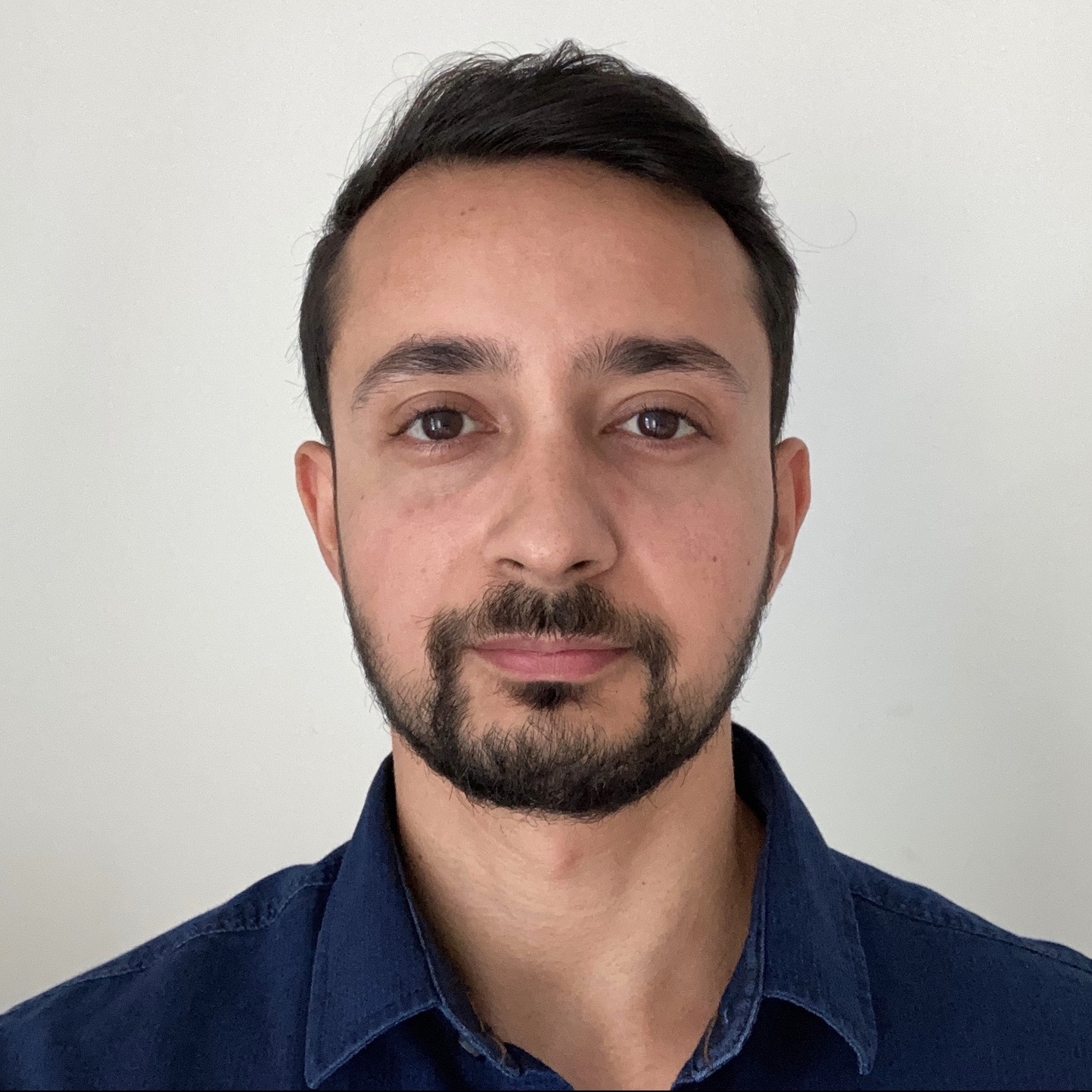}}]{Mehmet Aktukmak}
received the B.S. degree in electrical and electronics engineering from Hacettepe University in 2009, the M.S. degree in electrical and electronics engineering from Middle East Technical University in 2012, and the Ph.D. degree in electrical engineering from University of South Florida in 2020. He is currently working as postdoctoral research fellow in electrical and computer engineering department at the University of Michigan. His research interests include multimodal-multitask learning, Bayesian modelling, variational inference, and their applications to image/video processing, matrix completion, meta learning, and recommender systems.
\end{IEEEbiography}


\begin{IEEEbiography}
[{\includegraphics[width=1in,height=1.4in,clip,keepaspectratio]{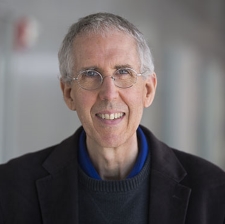}}]{Alfred O. Hero III}
received the B.S. (summa cum laude) from Boston University (1980) and the Ph.D from Princeton University (1984), both in Electrical Engineering. Since 1984 he has been with the University of Michigan, Ann Arbor, where he is the John H. Holland Distinguished University Professor of Electrical Engineering and Computer Science and the R. Jamison and Betty Williams Professor of Engineering. His primary appointment is in the Department of Electrical Engineering and Computer Science and he also has appointments, by courtesy, in the Department of Biomedical Engineering and the Department of Statistics. He is a Section Editor of the SIAM Journal on Mathematics of Data Science and a Senior Editor of the IEEE Journal on Selected Topics in Signal Processing . He is on the editorial board of the Harvard Data Science Review (HDSR) and serves as moderator for the Electrical Engineering and Systems Science category of the arXiv . He was co-General Chair of the 2019 IEEE International Symposium on Information Theory (ISIT) and the 1995 IEEE International Conference on Acoustics, Speech and Signal Processing.
He was founding Co-Director of the University’s Michigan Institute for Data Science (MIDAS) (2015-2018). From 2008-2013 he held the Digiteo Chaire d’Excellence at the Ecole Superieure d’Electricite, Gif-sur-Yvette, France. He is a Fellow of the Institute of Electrical and Electronics Engineers (IEEE) and the Society for Industrial and Applied Mathematics (SIAM). Several of his research articles have received best paper awards. Alfred Hero was awarded the University of Michigan Distinguished Faculty Achievement Award (2011), the Stephen S. Attwood Excellence in Engineering Award (2017), and the H. Scott Fogler Award for Professional Leadership and Service (2018). He received the IEEE Signal Processing Society Meritorious Service Award (1998), the IEEE Third Millenium Medal (2000), the IEEE Signal Processing Society Technical Achievement Award (2014), the Society Award from the IEEE Signal Processing Society (2015) and the Fourier Award from the IEEE (2020). Alfred Hero was President of the IEEE Signal Processing Society (2006-2008) and was on the IEEE Board of Directors (2009-2011) where he served as Director of Division IX (Signals and Applications). From 2011 to 2020 he was a member and Chair (2017-2020) of the Committee on Applied and Theoretical Statistics (CATS) of the US National Academies of Science.
Alfred Hero’s recent research interests are in high dimensional spatio-temporal data, multi-modal data integration, statistical signal processing, and machine learning. Of particular interest are applications to social networks, network security and forensics, computer vision, and personalized health. 
\end{IEEEbiography}

\end{document}